%% file: iclr2025_conference.tex
\documentclass{article} % For LaTeX2e
\usepackage{iclr2025_conference,times}

\input{math_commands.tex}

\usepackage{graphicx}
\newcommand{\phantomtitle}{\includegraphics[width=0.1\textwidth, trim=1in 4.5in 1in 4.5in]{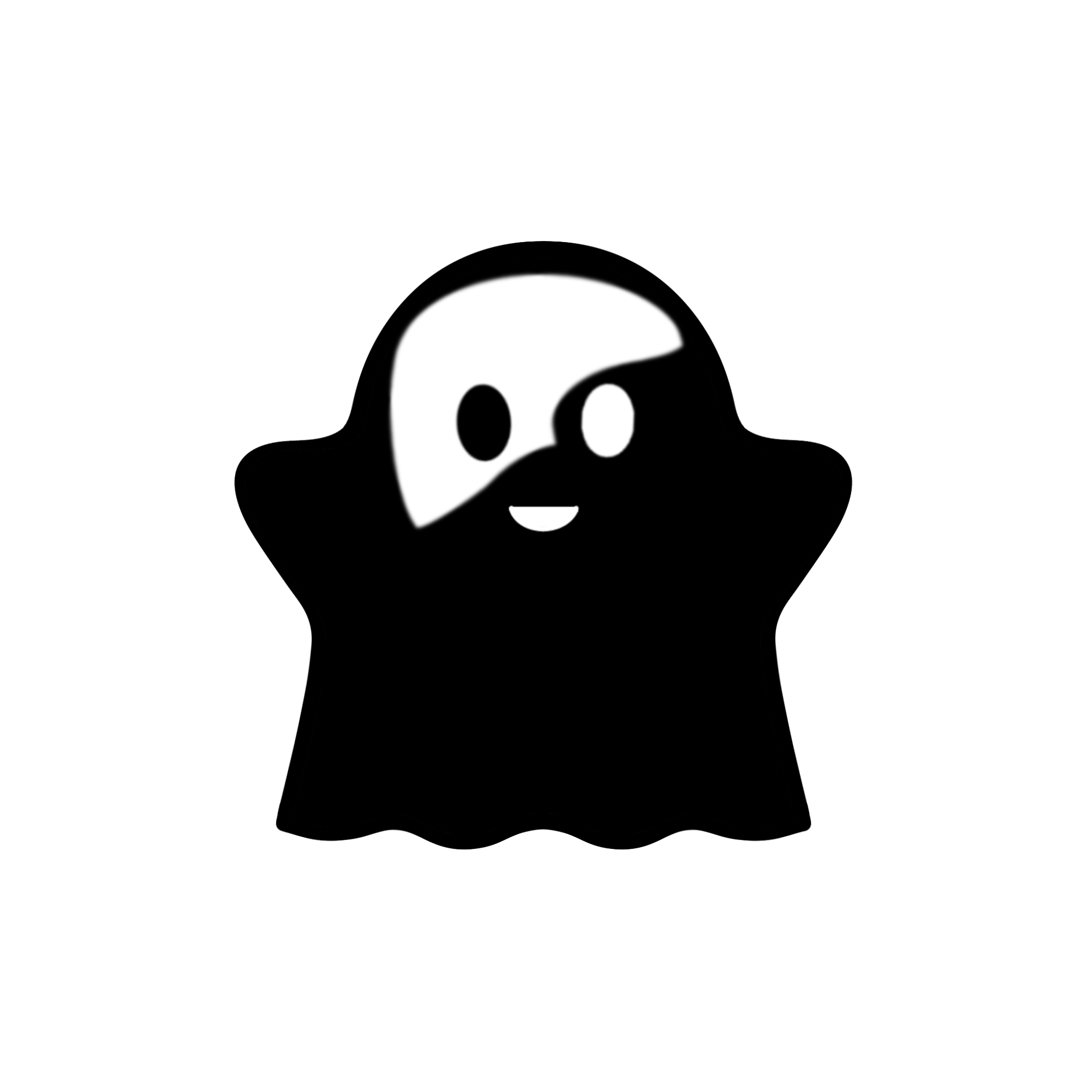}\hspace{0ex}} % phantom emoji
\newcommand{\aphantom}{\includegraphics[width=0.035\textwidth, trim=1in 4.0in 1in 4in]{figures/phantom_emoji.png}\hspace{0ex}} % phantom emoji

\usepackage{hyperref}
\usepackage{url}

\usepackage{float}
\usepackage{graphicx}
\usepackage{multirow}
\usepackage{amsmath}
\usepackage{xcolor} 
\usepackage{caption}
\usepackage{pifont} % c mark and x mark
\usepackage{color, colortbl} % for color table
\usepackage{arydshln} % for dash line in table
\usepackage{tabularx} % for matching width of table
\usepackage{footnote} % for footnote
\usepackage{booktabs} % for table

\title{\phantomtitle Phantom of Latent for Large Language and Vision Models}

\author{\hspace{4ex}Byung-Kwan Lee\\
\hspace{9ex}KAIST\\
\hspace{2ex}\texttt{\small leebk@kaist.ac.kr}
\And
\hspace{6ex}Sangyun Chung\\
\hspace{11ex}KAIST\\
\hspace{2.5ex}\texttt{\small  jelarum@kaist.ac.kr}
\And
\hspace{6ex}Chae Won Kim\\
\hspace{10.5ex}KAIST\\
\hspace{1ex}\texttt{\small  chaewonkim@kaist.ac.kr}
\And
\hspace{15ex}Beomchan Park\\
\hspace{19.5ex}KAIST\\
\hspace{10.5ex}\texttt{\small  bpark0810@kaist.ac.kr}
\And
\hspace{-3ex}Yong Man Ro\\
\hspace{0.5ex}KAIST\\
\hspace{-6ex}\texttt{\small  ymro@kaist.ac.kr}
}

\iclrfinalcopy % Uncomment for camera-ready version, but NOT for submission.
\begin{document}

\maketitle

\begin{abstract}
The success of visual instruction tuning has accelerated the development of large language and vision models (LLVMs). Following the scaling laws of instruction-tuned large language models (LLMs), LLVMs either have further increased their sizes, reaching 26B, 34B, and even 80B parameters. While this increase in model size has yielded significant performance gains, it demands substantially more hardware resources for both training and inference. Consequently, there naturally exists a strong need for efficient LLVMs that achieve the performance of larger models while being smaller in size. To achieve this need, we present a new efficient LLVM family with model sizes of 0.5B, 1.8B, 3.8B, and 7B parameters, \aphantom \textbf{Phantom}, which significantly enhances learning capabilities within limited structures. By temporarily increasing the latent hidden dimension during multi-head self-attention (MHSA), we make LLVMs prepare to look and understand much more vision-language knowledge on the latent, without substantially increasing physical model sizes. To maximize its advantage, we introduce \textbf{P}hantom \textbf{O}ptimization (\textbf{PO}) using both autoregressive supervised fine-tuning (SFT) and direct preference optimization (DPO)-like concept, which effectively follows correct answers while eliminating incorrect and ambiguous ones. \aphantom \textbf{Phantom} outperforms numerous larger open- and closed-source LLVMs, positioning itself as a leading solution in the landscape of efficient LLVMs. Code is available in \href{https://github.com/ByungKwanLee/Phantom}{https://github.com/ByungKwanLee/Phantom}.
\end{abstract}

%%%%%%%%%%%%%%%%%%%%%%%%%%%%%%%%%%%%%%%
% Introduction
\section{Introduction}
\label{sec:1}
In recent years, artificial general intelligence (AGI) has increasingly become a part of daily life, significantly enhancing our convenience. This trend is largely attributed to technical advancements of large language models (LLMs) and their impressive generalization performance, facilitated by instruction tuning~\citep{wei2022finetuned, chung2022scaling}. Building on this momentum, instruction tuning has expanded its realm into visual instruction tuning~\citep{liu2023visual}, integrating both language and vision as a format of text and image, under the use of pretrained LLMs. Based on them, numerous large language and vision models (LLVMs) have continuously emerged as multimodal LLMs and they have shown outstanding vision-language performances.

In terms of open-to-public regarding model architectures and their trained parameters, LLVMs can be categorized into open-source and closed-source models. For example, there are representative closed ones: GPT-4V~\citep{gptsyscard}, Gemini-Pro~\citep{team2023gemini}, and Qwen-VL-Plus~\citep{qwen, bai2023qwen}, all of which are renowned for their remarkable vision-language performances, large model sizes, and extensive number of dataset samples. In response, open-source LLVMs have tried to narrow the performance gap with their closed-source performances, by following the similar strategies the closed ones used, such as scaling up model sizes~\citep{liu2024llavanext, mckinzie2024mm1, li2024mini} (\textit{e.g.,} 26B, 34B, and 80B) and curating larger number of visual instruction tuning samples~\citep{hu2024mplug, fang2024vila, tong2024cambrian} (\textit{e.g.,} 4M, 6M, and 10M).

Along with them, several modules have focused on image-level understanding by leveraging numerous types of vision encoders~\citep{kar2024brave, lu2024deepseek, goncharova2024omnifusion, ranzinger2023radio, zhao2024cobra, li2024mini} and multiple computer vision models~\citep{chen2024spatialvlm, wang2024all, jiao2024enhancing, lee2024collavo, lee2024moai}. Additionally, a series of projectors have been employed alongside various vision encoders to improve fine-grained understanding~\citep{li2024mini, tong2024cambrian, ge2024convllava, chen2024evlm, yao2024minicpm} through partitioning the image. Besides, a multifaceted rationale-embedded projector~\citep{lee2024meteor} has been used to enhance real-world knowledge such as document, chart, and math.

% Figure 1
\input{figures/figure1}

However, these efforts — summarized as (a) scaling up model size, (b) curating larger datasets, and (c) incorporating additional modules and projectors — may not be regarded as a primary key to basically improve their own learning capabilities of LLVMs. In other words, there remains unexplored potential in fully utilizing LLVMs to align vision knowledge with language one and embed much more vision-language knowledge within limited structures, without relying on external modules and projectors. Beyond their limited learning capabilities, specifically, (a) and (b) bring in striking computational burdens during training, necessitating high-end GPUs with substantial VRAM. This (a) more becomes a critical drawback in devices with limited GPU resources, such as mobile phones and embedded boards. Furthermore, the high computational inference costs, associated with larger model sizes, exacerbate these issues, particularly for real-time applications such as augmented reality (AR) systems. As a result, deploying and operating LLVMs in such resource-constrained on-device environments becomes a major challenge.

To meet the two needs of maintaining model sizes while achieving superior performance, we present an efficient LLVM family, \aphantom \textbf{Phantom}, which stimulates enlarging vision-language learning capabilities within limited structures. When conducting multi-head self-attention (MHSA), \aphantom Phantom temporarily increases the latent hidden dimension and prepares to look and understand much more vision-language knowledge. Without significantly increasing the physical model size, we get an effect of increasing the dimension in query, key, and value, which we now call as Phantom Dimension. In order to maximally boost this advantage, we introduce \textit{\textbf{P}hantom \textbf{O}ptimization (\textbf{PO})}, inspired by RLHF and DPO~\citep{christiano2017deep, stiennon2020learning, ouyang2022training, rafailov2024direct, hong2024reference, meng2024simpo}. Unlike traditional preference-based methods, PO is designed to minimize the generation of incorrect and ambiguous answers. Since autoregressive supervised fine-tuning (SFT) primarily focuses on producing correct answers, PO provides \aphantom Phantom with additional guidance to avoid confusing answers by borrowing the recent DPO formulation~\citep{meng2024simpo}.

To do so, we first need a collection of incorrect and ambiguous answers. These are generated and filtered through GPT-4o(-mini) and human review from 2.8M visual instruction tuning samples covering diverse capabilities (details in Section \ref{sec:3}). This process resulted in the curation of 2M Phantom triples including question, its correct answer, and the corresponding incorrect and ambiguous answers (see Appendix~\ref{app:A}). By using the triple, \aphantom Phantom is trained with the two training steps, where we train vision projector and Phantom Dimension in the first step with the pretrained LLM frozen. In the second step, all components are trained together. Notably, PO utilizes SFT together with DPO-like concept throughout first training step, making \aphantom Phantom have an ability that follows correct answers while eliminating incorrect and ambiguous ones. In the experiment section, we demonstrate that handling the latent hidden dimension and using PO enhances vision-language performances by a large margin. As a result, we release an efficient LLVM family \aphantom Phantom with 0.5B, 1.8B, 3.8B, and 7B model sizes, which outperform open- and closed-source LLVMs, establishing a leading solution in the realm of efficient LLVMs.

% Figure 2
\input{figures/figure2}

Our contribution can be summarized into two main aspects:
\begin{itemize}
\item We present a new efficient large language and vision model (LLVM) Family, \aphantom \textbf{Phantom}, which temporarily increases the latent hidden dimension during multi-head self-attention (MHSA) to enhance vision-language learning capabilities within limited structures.
\item Curating efficient size 2M number of Phantom triples, we introduce a training strategy of \textbf{P}hantom \textbf{O}ptimization (\textbf{PO}) which avoids incorrect and ambiguous answers, showcasing more advancements across numerous evaluation benchmarks.
\end{itemize}
%%%%%%%%%%%%%%%%%%%%%%%%%%%%%%%%%%%%%%%

%%%%%%%%%%%%%%%%%%%%%%%%%%%%%%%%%%%%%%%
% Related Works
\section{Related Works}
\label{sec:2}

\paragraph{Large Language and Vision Models.}
To bridge the performance gap with closed-source LLVMs, open-source LLVMs have adopted three primary strategies: scaling up model size, curating larger datasets, and incorporating additional modules or projectors. For instance, LLaVA-NeXT ~\citep{liu2024llavanext}, MM1~\citep{mckinzie2024mm1}, Yi-VL~\citep{young2024yi} and MiniGemini~\citep{li2024mini} build model variants with parameters up to 34B. Concurrent to these efforts, mPLUG-Owl~\citep{hu2024mplug}, VILA$^2$~\citep{fang2024vila}, and Cambrian-1~\citep{tong2024cambrian} curate high-quality visual instruction tuning datasets specialized for diverse visual capabilities. Lastly, recent works have leveraged various vision encoders ~\citep{kar2024brave, lu2024deepseek, goncharova2024omnifusion, ranzinger2023radio, zhao2024cobra, li2024mini} and integrated external computer vision modules~\citep{chen2024spatialvlm, wang2024all, jiao2024enhancing, lee2024collavo, lee2024moai} to expand LLVMs' perception capabilities. Alongside using extra vision encoders, several works utilize projectors to extract hierarchical features of images~\citep{li2024mini, tong2024cambrian, ge2024convllava, chen2024evlm, yao2024minicpm} or to improve real-world knowledge comprehension such as document analysis, chart interpretation, and mathematical reasoning~\citep{lee2024meteor}.

While these approaches enhance downstream task performance, they do not address the core challenge of improving the intrinsic learning capabilities of LLVMs. Scaling up model size or employing larger instruction tuning datasets leads to substantial computational burdens. In addition, relying on extra visual encoders or computer vision modules brings in external visual knowledge, but they mainly focus on visual perception-related capabilities and their additional parameters may also lead the burden. This underscores the need for developing more efficient LLVMs with enhanced inherent capabilities that do not depend on such resource-intensive strategies.

\paragraph{Efficient Modeling.} 
In an effort to enhance the fundamental capabilities of LLMs while maintaining model size, several works for natural language processing has increasingly focused on developing smaller model sizes~\citep{thawakar2024mobillama, mehta2024openelm, liu2024mobilellm}, network pruning~\citep{ma2023llmpruner, men2024shortgpt, ashkboos2024slicegpt}, and quantization~\citep{li2023loftq, shao2024omniquant, park2024lutgemm}. These approaches primarily aim to accelerate training speed and reduce inference time while retaining performance, rather than boosting performances or improving LLVMs' embedding capabilities of vision-language knowledge within the limited structures. While efficient modeling has been extensively explored for LLMs, the design of efficient vision-language models (LLVMs) remains underexplored. A recent work, TroL~\citep{lee2024trol}, uniquely introduces a layer traversing technique that reuses layers in a token-wise manner to potentially embed more vision-language knowledge. However, it faces significant challenges, such as increased inference time due to doubling layer propagation and critical issues with key-value cache storage, preventing it from fully realizing its potential for efficient LLVMs.

In response to the need for efficient yet high-performing LLVMs, we introduce a new efficient LLVM family, \aphantom \textbf{Phantom}, which enhances the embedding capability of vision-language knowledge by temporarily increasing the latent hidden dimension during multi-head self-attention (MHSA). This innovation, combined with 2M Phantom triples to guide LLVMs towards correct answers while avoiding confusion, is expected to pave the way for more efficient LLVMs in both training and inference and to represent a crucial first step in advancing the field.

% Phantom of Latent
\section{\phantomtitle Phantom}
\label{sec:3}

\paragraph{Overview of Model Architecture.} As shown in Figure~\ref{fig:3}(a), the architecture of \aphantom Phantom model consists of vision encoder, vision projector, and a multimodal language model including word embedding and language model head, which follows a common configuration used in open-source LLVMs~\citep{liu2023visual, liu2023improved, bai2023qwen, chen2023sharegpt4v, mckinzie2024mm1}. Specifically, we utilize InternViT-300M~\citep{chen2023internvl} as the vision encoder instead of CLIP-L-428M~\citep{clip}, due to its superior ability to align text-to-image representations through contrastive learning with large language models (LLMs). The vision projector is constructed using two fully connected layers, where GELU~\citep{hendrycks2016gaussian} activation function is interleaved with each layer. For multimodal LLM component, we initialize it using pretrained LLMs across various sizes, selected for their state-of-the-art performance within their respective size: Qwen2-0.5B~\citep{yang2024qwen2}, InternLM2-1.8B~\citep{cai2024internlm2}, Phi3-mini-3.8B~\cite{abdin2024phi}, and InternLM2.5-7B~\citep{cai2024internlm2}.

\paragraph{Gathered Visual Instruction Tuning Sample Configuration.} To cover a broad range of capabilities, we compile 2.8M visual instruction tuning samples across multiple datasets, encompassing various domains such as fundamental image understanding, real-world common-sense knowledge, non-object visual concepts (e.g., documents, charts, diagrams, symbols, and signs), and general mathematical problems. Our dataset collection includes basic image understanding samples from ShareGPT4o-Images (57K)~\citep{sharegpt4o}, ShareGPT4V (755K)~\citep{chen2023sharegpt4v}, ALLaVA-VFLAN/Text (548K)~\citep{chen2024allava}, and MiniGemini (27K)~\citep{li2024mini} targeting tasks for DocVQA~\citep{mathew2021docvqa}, ChartQA~\citep{masry2022chartqa}, DVQA~\citep{kafle2018dvqa}, and AI2D~\citep{kembhavi2016diagram}. Additionally, we collect LLaVA-HD (116K)~\citep{zhang2024beyond} for Science and Mathematical Reasoning (SMR), supporting ArXivQA~\citep{li2024multimodal} and TextbookQA~\citep{kembhavi2017you}, and we further integrate document understanding samples from mPLUG-DocOwl1.5-Downstream/Reasoning (599K)~\citep{hu2024mplug} and general mathematical problems from GLLaVA (177K)~\citep{gao2023g}, MathVision (3K)~\citep{wang2024measuring}, MathInstruct (262K)~\citep{yue2023mammoth}, and MathPlus (304K)~\citep{yue2024mammoth2}.

\paragraph{Curation of Phantom Triples.} From the gathered 2.8M visual instruction tuning samples, we generate incorrect and ambiguous answers based on the existing question-answer pairs. To reduce data generation costs, we utilize GPT-4o-mini with the following prompt: \textit{``Question: \{\}. Answer: \{\}. Based on the question and the answer, make an incorrect and ambiguous answer compared to the original one. The length of the original answer should be maintained. Do not include any additional text.''}. Here, \{\} serves as a placeholder. Next, we employ GPT-4o to validate the generated responses using the prompt: \textit{``Original Answer : \{\}. Incorrect and Ambiguous Answer: \{\}. Provide `Yes' or `No', where `Yes' means it is incorrect and ambiguous answer compared to the original one, `No' means it is correct answer compared to the original one. Do not include any additional text.''}. All samples labeled `No' are discarded, while the `Yes'-labeled samples undergo human review to verify if they are genuinely confusing. Through this process, we curate 2M Phantom Triples, consisting of a question, its correct answer, and a corresponding confusing answer.

% figure 3
\input{figures/figure3}

\paragraph{Realization of Phantom Dimension.}
 For better understanding, Figure~\ref{fig:3}(b) represents the simple overview of how Phantom Dimension works. We utilize start of sequence (sos) token that will serve as a key in enhancing the latent hidden dimension for the query, key, and value components in multi-head self-attention (MHSA) layers. The latent feature on the location of sos token is propagated into QKV linear function, and we denote its outputs as $Q^{*}_{l}\in\mathbb{R}^{d_q}$, $K^{*}_{l}\in\mathbb{R}^{d_{kv}}$, and $V^{*}_{l}\in\mathbb{R}^{d_{kv}}$ at each layer $l$. Note that $d$ denotes the latent hidden dimension. $Q^{*}_{l}$, $K^{*}_{l}$, and $V^{*}_{l}\in\mathbb{R}^{d_{kv}}$ are supposed to inject into the multi-head cross-attention (MHCA) module. A natural question arises: \textit{Why inject these features into the cross-attention module?} The reason lies in the dynamic length $N$ of user input tokens, which varies with the question length. Therefore, these features need to have dimension $Q^{*}_{l}\in\mathbb{R}^{N\times d_{q}}$, $K^{*}_{l}\in\mathbb{R}^{N\times d_{kv}}$, and $V^{*}_{l}\in\mathbb{R}^{N\times d_{kv}}$ since sos token only represents a single token. Therefore, it must be expanded to match the $N$ tokens of the input sequence, and the cross-attention module make these features expanded into input sequence token number $N$, as follows:
\begin{equation}
    \begin{split}
    Q^{*}_{l} &\leftarrow \text{MHCA}(q=Q_{l}, k/v=Q^*_l),\\
    K^{*}_{l} &\leftarrow \text{MHCA}(q=K_{l}, k/v=K^*_l),\\
    V^{*}_{l} &\leftarrow \text{MHCA}(q=V_{l}, k/v=V^*_l),
    \end{split}
\end{equation}
where we change their dimension into $Q_l$: $\mathbb{R}^{N\times h_q\times \frac{d_q}{h_q}}$ and $K_l, V_l$: $\mathbb{R}^{h_{kv}\times \frac{d_{kv}}{h_{kv}}}$ for conducting multi-head cross attention with head number $h_q$ and $h_{kv}$. Next, in order to make LLVMs embed much more vision-language knowledge, we enlarge the latent hidden dimension by concatenating the original query, key, and value matrices with the cross-attended outputs dimension-wise, yielding $\begin{bmatrix} Q_{l} & Q^{*}_{l} \end{bmatrix} \in\mathbb{R}^{N\times h_q\times \frac{2d_q}{h_q}}$, $\begin{bmatrix} K_{l} & K^{*}_{l} \end{bmatrix} \in\mathbb{R}^{N\times h_{kv}\times \frac{2d_{kv}}{h_{kv}}}$, and $\begin{bmatrix} V_{l} & V^{*}_{l} \end{bmatrix} \in\mathbb{R}^{N\times h_{kv}\times \frac{2d_{kv}}{h_{kv}}}$. We then apply multi-head self-attention (MHSA) used in multimodal LLM to these concatenated ones:
\begin{equation}
    O_l = \text{Softmax}\left(\lambda\left(\frac{2d_q}{h_q}\right)^{-\frac{1}{2}}\begin{bmatrix} Q_l & Q^{*}_l \end{bmatrix} \begin{bmatrix} K_{l} & K^{*}_{l} \end{bmatrix}^\top \right) \begin{bmatrix} V_l & V^{*}_l\end{bmatrix},
\end{equation}
where $\lambda$ denotes a regularization parameter, and $O_l\in\mathbb{R}^{N\times h_q\times \frac{2d_q}{h_q}}$ represents the output features of MHSA. After its computation, the output features should return to the original hidden dimension, as they will be propagated through the remaining transformer modules, such as feed-forward network (FFN). At this stage, we aim to compress the output features while minimizing information loss as much as possible. To achieve this, we split the output $O_l$ into two halves: $O_l$[:, :, :$\frac{d_q}{h_q}$] and $O_l$[:, :, $\frac{d_q}{h_q}$:] (Python slicing format), denoted as $\bar{O_{l}}\in\mathbb{R}^{N\times h_q\times \frac{d_q}{h_q}}$ and $\tilde{O_{l}}\in\mathbb{R}^{N\times h_q\times \frac{d_q}{h_q}}$, respectively. To flexibly mix them, weighted-average operation is employed, and then finally we can get the compressed outputs $O_l\gets \bar{w} \odot \bar{O_{l}} + \tilde{w} \odot \tilde{O_{l}}$ where $\odot$ is element-wise multiplication, and
\begin{equation}
    \bar{w} = \frac{e^{f(\bar{O_l})}}{e^{f(\bar{O_l})}+e^{g(\tilde{O_l})}},\quad \tilde{w} = \frac{e^{f(\tilde{O_l})}}{e^{f(\bar{O_l})}+e^{g(\tilde{O_l})}},
\end{equation}
where $f$ and $g$ comprise each one fully-connected layer: $\mathbb{R}^{N\times h_q\times\frac{d_q}{h_q}}\rightarrow\mathbb{R}^{N\times h_q}$, and the compressed outputs are then propagated into remaining modules with root mean square (RMS) layer normalization~\citep{ba2016layernormalization, zhang2019root} and Add\&Norm operation.

% Table 1
\input{tables/table1}

\paragraph{Implementation of Phantom Optimization.} To fully leverage the enhanced learning capability provided by Phantom Dimension, we introduce Phantom Optimization (PO), which is heavily inspired by Direct Preference Optimization (DPO)~\citep{rafailov2024direct}. While methods such as RLHF~\citep{christiano2017deep} and DPO are designed to optimize towards human or AI-driven preferences, PO is tailored to follow correct answer and reduce incorrect and ambiguous answers during training. To reduce the computational complexity of incorporating an additional reference model, we adopt the loss formulation from SimPO~\citep{meng2024simpo}. Similar to ORPO~\citep{hong2024reference}, we simultaneously use autoregressive supervised fine-tuning (SFT). This enables \aphantom Phantom to effectively reinforce correct answers $y^{+}$ while eliminating incorrect and ambiguous ones $y^{-}$ in response to a given prompt $x$. This formulation can be expressed as follows:
\begin{equation}
    \label{eqn:po}
    \min_{\theta}\mathcal{L}_{\text{PO}} = \mathcal{L}_{\text{SFT}} -\E_{\mathcal{D}}\left[\log\sigma\left(\frac{\beta}{|y^+|}\log{\pi_\theta(y^{+}|x)} - \frac{\beta}{|y^-|}\log{\pi_\theta(y^{-}|x)} - \gamma\right)\right],
\end{equation}
where $\theta$ represents the trainable parameters and $\mathcal{L}_{\text{SFT}}$ denotes the supervised fine-tuning loss for question-answer pairs. We implement a two-step training strategy. In the first step, which focuses on vision and language alignment, the parameters of the pretrained LLM are frozen. We then train the parameters of vision projector and the components related to Phantom Dimension (MHCA and the functions $f$ and $g$). In the second step, we unfreeze all parameters and train them all at once. We apply PO throughout the first training step only, not to interrupt multimodal LLM's own text generation ability because the positive and negative answers $y^{+}/y^{-}$ are mostly generated by closed-source LLVMs instead of instruction fine-tuned self model, which is totally different strategy from RLHF and DPO. For verification, we show the performance degradation in experiment section when using PO in the second training step. 
%%%%%%%%%%%%%%%%%%%%%%%%%%%%%%%%%%%%%%%

%%%%%%%%%%%%%%%%%%%%%%%%%%%%%%%%%%%%%%%
% Experiments
\section{Experiments}
\label{sec:4}

% Table 2
\input{tables/table2}

\paragraph{Implementation Details.} To ensure successful reproducibility, we outline four key technical aspects of \aphantom Phantom: (a) the detailed architecture of the backbone multimodal LLMs, vision encoder, and vision projector, (b) the structure of the multi-head cross-attention (MHCA) module in Phantom Dimension, (c) the computing environments and bit quantization configurations, and (d) the procedures for training and inference.

\paragraph{(a)} We utilize Qwen2~\citep{yang2024qwen2}, Phi-3-mini~\citep{abdin2024phi}, and InternLM2/2.5~\citep{cai2024internlm2} as the backbone multimodal LLMs. Specifically, Qwen2-0.5B is configured with $h_q=14$, $h_{kv}=2$, a hidden dimension of $d_q=896$, and 24 layers; InternLM2-1.8B with $h_q=16$, $h_{kv}=8$, a hidden dimension of $d_q=2048$, and 24 layers; Phi-3-mini-3.8B with $h_q=32$, $h_{kv}=32$, a hidden dimension of $d_q=3072$, and 32 layers; and InternLM2.5-7B with $h_q=32$, $h_{kv}=8$, a hidden dimension of $d_q=4096$, and 32 layers. For the vision encoder, we employ InternViT-300M~\citep{chen2023internvl}, which has a hidden dimension of 1024 and 24 layers. The vision projector is designed as MLP that adjusts the hidden dimension from 1024 to match the corresponding multimodal LLM's latent hidden dimension.

\paragraph{(b)} In each layer, MHCA consists of four linear modules for the query, key, value, and output of the multi-head self-attention operation, where MHCA has similar head dimension for MHSA. For the 0.5B model, the number of parameters required for MHCA module is approximately 1.2M, calculated as $(\frac{896\text{ (hidden dimension)}}{14\text{ (number of heads)}})^2 \times 4 \text{ (linear modules)} \times 24 \text{ (layers)} \times 3 \text{ (}qkv\text{)}$. Similarly, the required parameters for the 1.8B, 3.8B, and 7B models are 4.8M, 3.7M, and 6.2M, respectively. These additional parameters do not significantly impact the overall model size compared with 0.5B, 1.8B, 3.8B, and 7B. Note that, the regularization parameter $\lambda$ during MHSA is set to $\sqrt{2}$.

% Table 3
\input{tables/table3}

\paragraph{(c)} In a computing environment utilizing 8$\times$NVIDIA RTX A6000 48GB GPUs and 8$\times$NVIDIA RTX 3090 24GB GPUs, \aphantom Phantom's training and inference processes take place. To conduct efficient training, each step undergoes a single epoch of training using 8-bit quantization and bfloat16 data format~\citep{kalamkar2019study} for every backbone multimodal LLM. Following bit quantization, we apply QLoRA~\citep{hu2021lora, dettmers2023qlora} to both vision encoders and backbone multimodal LLMs across all linear layers, using 256 rank and 256 alpha parameters.

\paragraph{(d)} For Phantom Optimization, we choose equal hyperparameters used in SimPO~\citep{meng2024simpo}: $\beta=2$ and $\gamma=0.5$. For training, AdamW optimizer~\citep{loshchilov2018decoupled} is applied, and cosine annealing adjusts the learning rate from 1e-5 to 1e-6 throughout each training step. For multimodal LLM, gradient checkpointing~\citep{sohoni2019low} is employed to manage memory efficiently. A gradient accumulation of 4 leads to batch sizes totaling 128 for each training step, with each step taking roughly two to five days depending on model size. For inference efficiency, \aphantom Phantom is validated using the same quantization level in training, and we make Phantom Dimension cache: $Q^{*}_l$, $K^{*}_l$, and $V^{*}_l$ in each layer to get speedy inference like kv-cache technique, where we use deterministic beam search~\citep{freitag-al-onaizan-2017-beam} ($n=3$). Memory-efficient scaled dot product attention (SDPA) and FlashAttention2~\citep{dao2022flashattention, dao2023flashattention} accelerates multi-head self-attention (MHSA) computation for Phantom Dimension, benefiting from its hardware-aware ability to mitigate the overhead from the increased latent hidden dimension. 

% Table 4
\input{tables/table4}

\paragraph{Validation and Ablation Studies.} We present an overview of \aphantom Phantom's vision-language performance in Figure~\ref{fig:1}-\ref{fig:2}, and evaluate it on generally used standard evaluation benchmarks as shown in Table~\ref{tab:1}-\ref{tab:2}. In the table, LLaVA-OneVision-8B~\citep{li2024llava} uses significant number of image tokens up to 7290 with three training steps on 558K+4M+3.2M datasets. To highlight the benefits of \aphantom Phantom, Table~\ref{tab:3} reports performance on more challenging evaluation benchmarks. These results demonstrate that \aphantom Phantom offers a significant advantage on tasks requiring reasoning abilities and densely learned knowledge.  Descriptions of the evaluation benchmarks can be found in Appendix~\ref{app:B}, and \aphantom Phantom's text generation quality is illuminated in Appendix~\ref{app:C}. In conclusion, \aphantom Phantom achieves outstanding performance across numerous vision-language tasks, with a large margin over competing LLVMs, despite having a smaller model size and fewer instruction tuning samples. To better understand the source of this effectiveness, Table~\ref{tab:4} presents an ablation study focusing on three key factors: (a) Weighted-Average (WA), (b) Phantom Dimension (PD), and (c) Phantom Optimization (PO). The results reveal several insights: (1) PD significantly enhances vision-language performance, as increasing the latent hidden dimension improves the embedding of vision-language knowledge; (2) WA is more effective than simple summation or averaging for compressing output features; and (3) PO yields greater performance gains when combined with PD and when applied only during the first training step with a frozen pretrained LLM. Besides, we investigated the effect of replacing the sos token with alternative tokens. We observed using the token that appears earlier in the user question prompt, before the question, is more effective. Regarding inference speed, we measured computation time and found only a marginal 10\% difference in tokens-per-second between the settings with and without PD. It is definitely attributed to hardware-level computed operation using SDPA and FlashAttention2~\citep{dao2022flashattention, dao2023flashattention}.

\paragraph{Discussion and Limitation.} The development of high-performing LLVMs increasingly depends on combining diverse models~\citep{lu2024deepseek, lee2024collavo, lee2024moai, lee2024meteor, zong2024mova, shi2024eagle} and refining existing architectures~\citep{liu2024mobilellm, lee2024trol}, as many aspects of these systems remain unexplored. However, such structural modifications often leads to substantial low-level programming when addressing both development and production-level demands. In response, we will do comprehensive exploration of significantly larger open-source LLVMs, without additional architectural changes. Although there has been a growing trend toward open-source LLVMs, much of the research continues to focus on closed-source LLVMs such as GPT-4V and Gemini-Pro. We either had used GPT-4o-mini and GPT-4o. Therefore, we believe there is untapped potential not only in utilizing the textual outputs of larger open-source LLVMs but also in accessing deeper insights, such as layer-wise features or full parameter sets across layers. Moving forward, we plan to investigate layer-wise distillation methods, which go beyond traditional distillation, to transfer knowledge into models with entirely different architectures using human-understandable language. This direction promises to open up exciting possibilities in a more easier way to develop efficient LLVMs, such as transferring knowledge across heterogeneous structures.

%%%%%%%%%%%%%%%%%%%%%%%%%%%%%%%%%%%%%%%

%%%%%%%%%%%%%%%%%%%%%%%%%%%%%%%%%%%%%%%
% Conclusion
\section{Conclusion}
\label{sec:5}
We present an efficient LLVM family \aphantom Phantom with significantly enhanced learning capabilities within limited model sizes. By introducing Phantom Optimization (PO) that leverages both autoregressive supervised fine-tuning (SFT) and DPO-like concept, it effectively learns and boosts vision-language performances. Remarkably, despite being smaller than many high-performing LLVMs with larger model sizes, \aphantom Phantom demonstrates comparable or even superior performance, making it a promising solution for resource-constrained environments. Our results underscore the power of latent space optimization in boosting both efficiency and performance, offering a pathway toward more efficient LLVMs for various applications.
%%%%%%%%%%%%%%%%%%%%%%%%%%%%%%%%%%%%%%%

%%%%%%%%%%%%%%%%%%%%%%%%%%%%%%%%%%%%%%%
% References
\clearpage
\bibliography{iclr2025_conference}
\bibliographystyle{iclr2025_conference}
%%%%%%%%%%%%%%%%%%%%%%%%%%%%%%%%%%%%%%%

%%%%%%%%%%%%%%%%%%%%%%%%%%%%%%%%%%%%%%%
% Appendix
\appendix

% Appendix A
\clearpage
\section{Phantom Triples}
\label{app:A}
\input{appendix/AppendixA}

% Appendix B
\clearpage
\section{Description of Evaluation Benchmarks}
\label{app:B}
\begin{itemize}    
    \item \textbf{SQA-IMG (SQA$^{\text{I}}$)}~\citep{lu2022learn} is part of the broader ScienceQA (SQA) dataset, which aims to improve reasoning and interpretability in AI systems through science-based question answering. This dataset covers a wide range of science disciplines, featuring 26 different topics in natural, social, and language sciences, all accompanied by annotated answers, lectures, and explanations. SQA-IMG includes image-related samples, amounting to 10,332 question-answer pairs.
    
    \item \textbf{AI2D}~\citep{kembhavi2016diagram} or AI2 Diagrams, addresses diagram interpretation and reasoning challenges, focusing on syntactic parsing and semantic understanding. It supports research into diagram structure and element relationships, critical for tasks like diagram-based question answering. This collection includes over 5,000 diagrams from elementary science topics, along with over 15,000 multiple-choice questions.
    
    \item \textbf{ChartQA}~\citep{masry2022chartqa} develops to challenge and improve question answering systems that deal with data visualizations like bar charts, line charts, and pie charts. This dataset tests systems on questions requiring arithmetic and logical reasoning and includes both human-generated and machine-created question-answer pairs. It comprises 32,719 samples in total.
    
    \item \textbf{SEED-IMG (SEED$^{\text{I}}$)}~\citep{li2023seed}, a subset of SEED-Bench, evaluates the generative comprehension skills of multimodal large language models (MLLMs) with a focus on spatial and temporal understanding. It offers several subsets mapped to 12 evaluation dimensions across image and video modalities, with SEED-IMG specifically concentrating on images.

    \item \textbf{SEED-Bench-2-Plus}~\citep{li2024seed} evaluates multimodal large language models in their ability to understand text-rich visual content, common in real-world settings like charts, maps, and website interfaces. This benchmark specifically measures how effectively MLLMs can interpret these complex, text-rich scenarios that require simultaneous comprehension of visual and textual information. The benchmark is divided into three main categories—Charts, Maps, and Webs, and further subdivided into 63 unique data types with 2.3k multiple-choice questions.
    
    \item \textbf{POPE}~\citep{li2023evaluating} introduces a method to systematically assess the tendency of LLVMs to falsely generate nonexistent objects in images. This method turns the evaluation into a binary classification task using polling questions, providing a fair and adaptable approach.
    
    \item \textbf{HallusionBench (HallB)}~\citep{liu2023hallusionbench} is crafted to evaluate and explore visual illusions and knowledge hallucinations in large language and vision models (LLVMs). This benchmark uses carefully crafted example pairs to identify model failures, featuring diverse visual-question pairs including subsets focused on illusions, math, charts, tables, maps, and OCR. It includes 346 images and 1,129 questions.
    
    \item \textbf{MME}~\citep{fu2023mme} serves as a comprehensive evaluation framework for Multimodal Large Language Models (MLLMs), focusing on various perception and cognition tasks through 14 sub-tasks like coarse and fine-grained recognition, OCR, and commonsense reasoning. This benchmark aims to address existing evaluation gaps and ensures a thorough testing environment for MLLMs.
    
    \item \textbf{MathVista}~\citep{lu2023mathvista} is an extensive benchmark designed to test visual-based mathematical reasoning in AI models. It integrates visual understanding in evaluating models' abilities to solve math problems that involve visuals. The dataset consists of three subsets: IQTest, FunctionQA, and PaperQA, totaling 6,141 examples.
    
    \item \textbf{MMB, MMB-Chinese (MMB$^{\text{CN}}$)}~\citep{liu2023mmbench} aims to establish a robust evaluation standard for vision language models by covering a broad spectrum of necessary multimodal comprehension skills (20 fine-grained abilities) in both English and Chinese. This benchmark consists of 3,217 questions gathered from various sources to challenge different facets of LLVMs.
    
    \item \textbf{MM-Vet}~\citep{yu2023mm} is designed to systematically evaluate LMMs on complex tasks requiring multiple vision language (VL) capabilities. It tests recognition, knowledge, OCR, spatial awareness, language generation, and math, integrating these abilities into 16 different task combinations. The dataset includes 200 images and 218 questions, each requiring the integration of multiple capabilities.

    \item \textbf{MM-Vet-v2}~\citep{yu2024mm} evaluates a wide range of integrated abilities in large multimodal models, such as Recognition, Knowledge, Optical Character Recognition (OCR), Spatial Awareness, Language Generation, Math, and Image-Text Sequence Understanding. This version builds upon the original MM-Vet benchmark by adding tasks that involve comprehending sequential information from both images and text, which is essential for real-world scenarios. MM-Vet-v2 places a strong focus on assessing the model's capacity to interpret and reason through intricate image-text sequences. The benchmark includes 517 evaluation samples, a notable increase from the 217 samples in the original MM-Vet.
    
    \item \textbf{LLaVA Bench in the Wild(er) (LLaVA$^\text{W}$ and LLaVA-Wilder )}~\citep{liu2023visual, zhang2024lmms} assesses large multimodal models (LMM) on complex tasks and new domains through a collection of 24 images with 60 questions for `wild' and its more advanced version of `wilder'. This dataset features diverse settings, including indoor, outdoor, artworks, and memes, with each image accompanied by detailed descriptions and curated questions.
    
    \item \textbf{MMStar}~\citep{chen2024we} is crafted to precisely evaluate the true multimodal capabilities of LLVMs by ensuring that each sample critically relies on visual content for accurate answers while minimizing data leakage. It comprises 1,500 meticulously selected samples and is organized into six primary sub-benchmarks as follows:
        \begin{itemize}
        \item \textbf{Coarse perception (CP)}, which pertains to the ability to grasp and interpret the overarching features and themes of an image without focusing on minute details,
        \item \textbf{Fine-grained perception (FP)}, which denotes a detailed level of image comprehension that emphasizes the intricate and nuanced aspects of visual content,
        \item \textbf{Instance reasoning (IR)}, which encompasses advanced cognitive abilities aimed at understanding and interpreting individual and collective object attributes and their interrelations within an image,
        \item \textbf{Logical reasoning (LR)}, which involves a sophisticated framework of cognitive processes designed to interpret, deduce, and infer conclusions from visual content through a structured approach to logic and reasoning,
        \item \textbf{Science \& technology (ST)}, which includes a comprehensive framework for the application and integration of knowledge across a wide range of scientific and technological domains,
        \item \textbf{Math (MA)}, which is a fundamental pillar of logical and analytical reasoning and includes a broad spectrum of skills essential for understanding, applying, and interpreting quantitative and spatial information.
        \end{itemize}
    
    \item \textbf{MathVerse}~\citep{zhang2024mathverse} assesses the capabilities of Multi-modal Large Language Models (MLLMs) in visual mathematical reasoning, particularly their ability to understand visual diagrams and mathematical expressions. This dataset is categorized into three primary areas: plane geometry, solid geometry, and functions, and further detailed into twelve types like length and area, encompassing 2,612 visual mathematical challenges.

    To investigate how MLLMs process visual diagrams in mathematical reasoning, the creators of MathVerse developed six distinct versions of each problem, each version presenting different levels of multi-modal information. They initially established three specific classifications for the text content within the problems:
    \begin{itemize}
        \item \textit{Descriptive Information}, which includes content that is directly visible and explicitly depicted in the diagrams,
        \item \textit{Implicit Property}, which encompasses details that demand a more advanced visual perception yet less mathematical knowledge to interpret from the diagram,
        \item \textit{Essential Condition}, which pertains to crucial numerical or algebraic data necessary for solving the problem that cannot be inferred solely from the visual diagram.
    \end{itemize}

    Based on these categories, to thoroughly assess the true visual understanding capabilities of MLLMs and their utility in multi-modal mathematical contexts, the researchers created six versions or sub-benchmarks of each problem in MathVerse, described as follows:
    \begin{itemize}
        \item \textbf{Text dominant (TD)} version, which preserves all textual elements, including the three textual categories and the main question, prompting MLLMs to primarily depend on textual information.
        \item \textbf{Text lite (TL)} version reduces the \textit{Descriptive Information} from the Text dominant version, promoting reliance on the diagram for elementary data.
        \item \textbf{Text only (TO)} version removes the visual elements entirely, focusing on textual content to discern where MLLMs predominantly derive contextual information for problem solving.
        \item \textbf{Vision intensive (VI)} further excludes \textit{Implicit Property} from the Text lite version, urging MLLMs to intensify their visual analysis to gather essential cues for mathematical reasoning.
        \item \textbf{Vision dominant (VD)}, evolving from the Text lite version, omits \textit{Essential Condition} from the textual information and instead visually annotates these details in diagrams, compelling MLLMs to identify and accurately link these essential conditions solely through visual examination.
        \item \textbf{Vision only (VO)} eliminates all textual descriptions, presenting the problem exclusively through visual means and challenging MLLMs to decode and identify mathematical queries purely based on visual data, serving as the ultimate test of their visual reasoning skills in mathematics.
    \end{itemize}

    \item \textbf{VisualWebBench}~\citep{liu2024visualwebbench} assesses the capabilities of multimodal large language models (MLLMs) specifically in the web domain. It is designed to address the lack of a comprehensive benchmark that evaluates the unique characteristics of web pages and measures fine-grained abilities such as OCR, understanding, and grounding (Grd) in text-rich and interactive web environments. It covers a wide range of domains, including science, travel, sports, engineering, and government, and tasks such as captioning (Cap), WebQA (QA), heading OCR, element grounding (Grd), and action prediction (Pred), containing a total of 1,534 instances.

    \item \textbf{CV-Bench}~\citep{tong2024cambrian} is designed for vision-focused evaluation in multimodal large language models. This benchmark aims to fill the gaps in traditional benchmarks, which often fall short in thoroughly assessing visual grounding in real-world contexts. CV-Bench assesses the model's abilities in both 2D and 3D visual tasks using natural language questions. The evaluation is split into 2D tasks (such as spatial relationships and object counting) and 3D tasks (like depth order and relative distance), providing a well-rounded test of the model's visual comprehension with 2,638 carefully inspected examples.

    \item \textbf{BLINK}~\citep{fu2024blink} is created to assess the visual perception capabilities of multimodal large language models. It features 14 key visual perception tasks, which are based on traditional computer vision challenges but restructured into 3,807 multiple-choice questions that involve one or more images. These tasks address difficulties such as relative depth estimation, visual correspondence, forensic detection, and multi-view reasoning.
\end{itemize}

Additionally, we will continue to explore more challenging evaluation benchmarks to uncover previously unaddressed issues such as \citet{yu2024spark}, advancing \aphantom Phantom through ongoing technical development. By leveraging a wide range of methods established over the years~\citep{lee2020training, lee2020towards, kim2021distilling, lee2022masking, kim2023demystifying, lee2023mitigating, kim2023causal, kim2023mitigating, park2024robust, park2024integrating, kim2024improving}, we aim to drive innovative breakthroughs across both general and specialized tasks.

% Appendix C
\clearpage
\section{\phantomtitle Phantom Generation Quality}
\label{app:C}
\input{appendix/AppendixC}
%%%%%%%%%%%%%%%%%%%%%%%%%%%%%%%%%%%%%%%

\end{document}

%% file: math_commands.tex
\usepackage{amsmath,amsfonts,bm}

\def\eqref#1{equation~\ref{#1}}
\def\1{\bm{1}}

\DeclareMathAlphabet{\mathsfit}{\encodingdefault}{\sfdefault}{m}{sl}
\SetMathAlphabet{\mathsfit}{bold}{\encodingdefault}{\sfdefault}{bx}{n}

\newcommand{\E}{\mathbb{E}}

%% file: figures/figure1.tex
\begin{figure}[t!]
    \centering
    \vspace{-10mm}
    \includegraphics[width=0.9\textwidth]{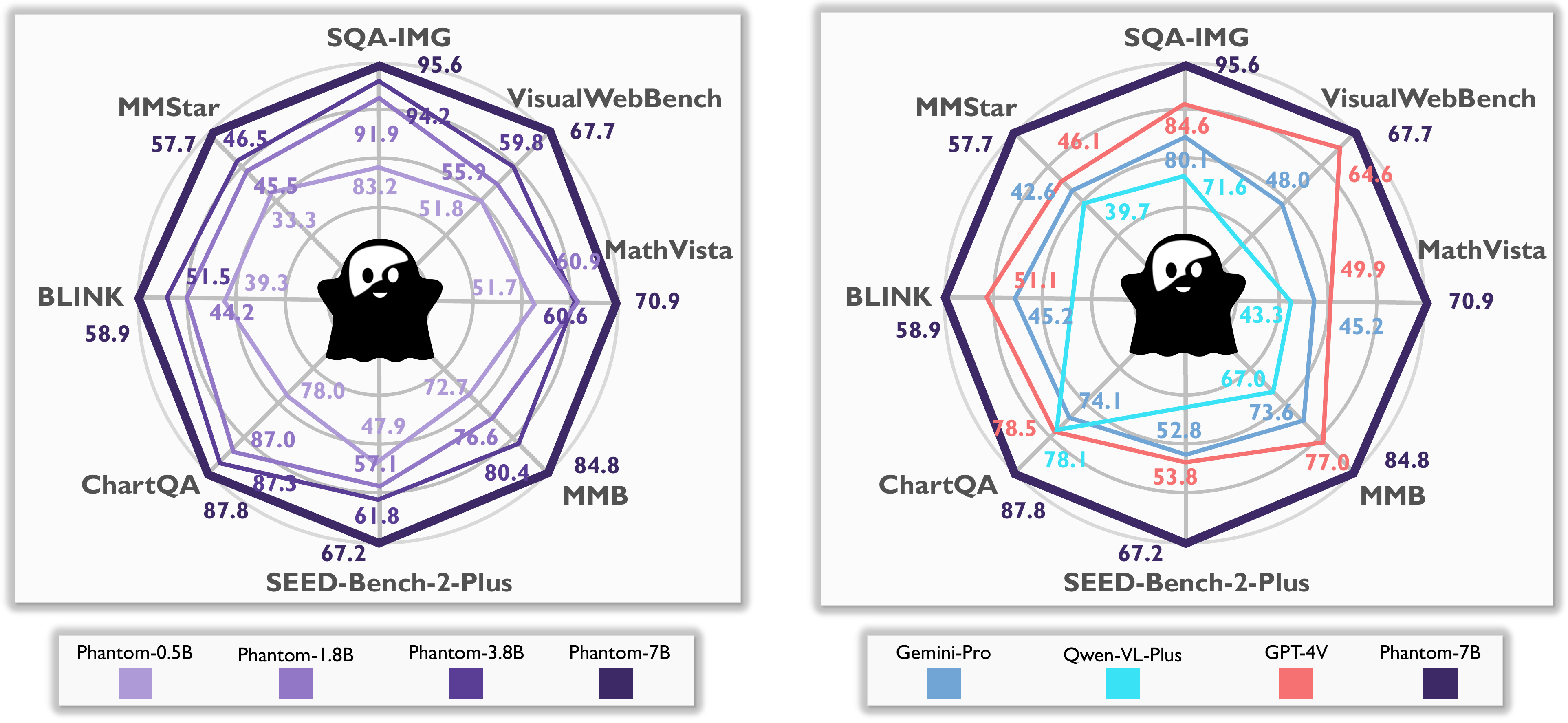}
    \vspace{-3mm}
    \caption{Overview of performances compared with \aphantom Phantom and closed-source LLVMs}
    \label{fig:1}
    \vspace{-6mm}
\end{figure}

%% file: figures/figure2.tex
\begin{figure}[t!]
\vspace{-10mm}
    \centering
    {
    \begin{minipage}[t]{0.525\linewidth}
    \includegraphics[width=\textwidth]{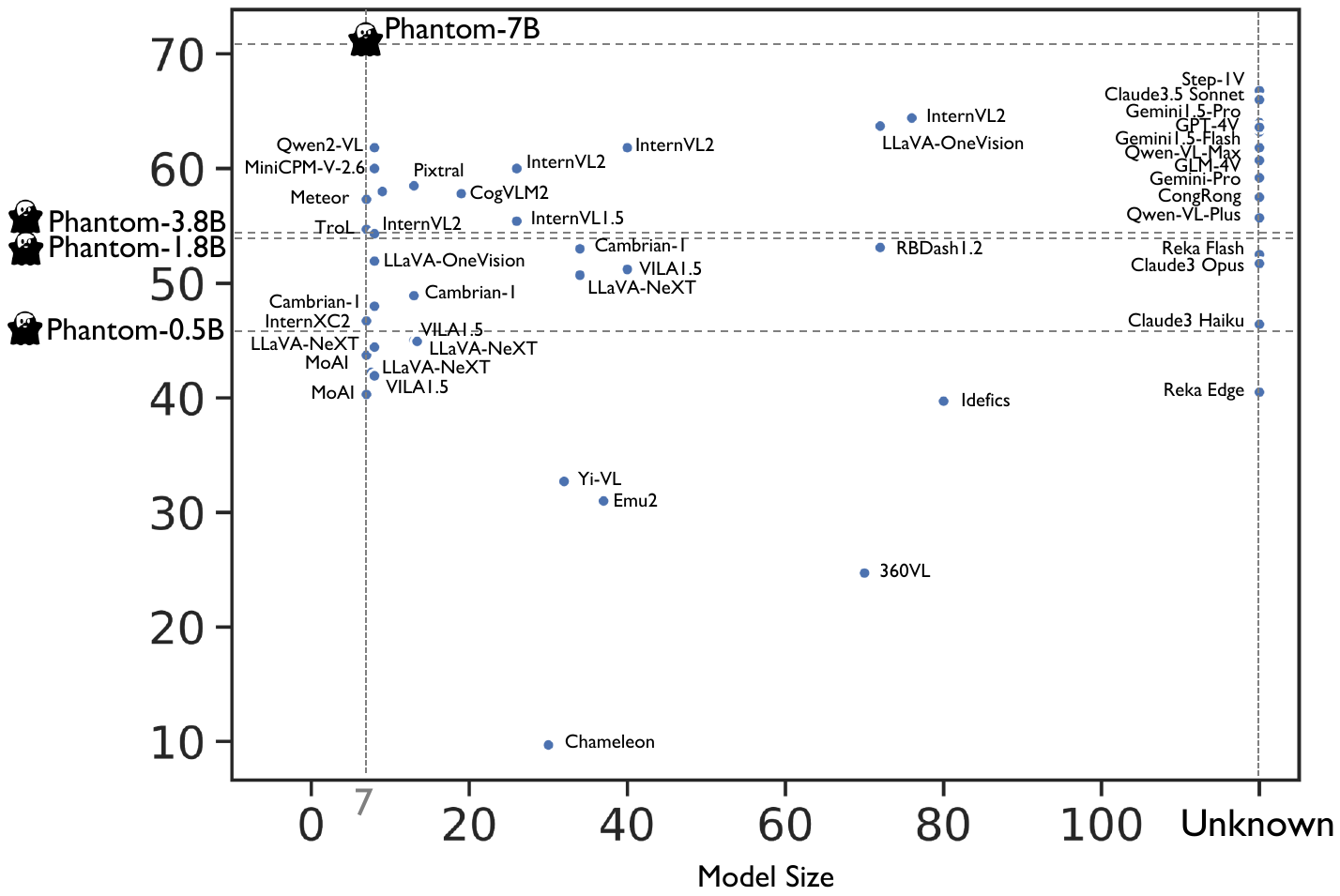}
    \vspace*{-6mm}
    \caption*{\hspace{10mm}(a) 7B$\sim$80B, and Unknown}
    \end{minipage}
    \begin{minipage}[t]{0.465\linewidth}
    \includegraphics[width=\textwidth]{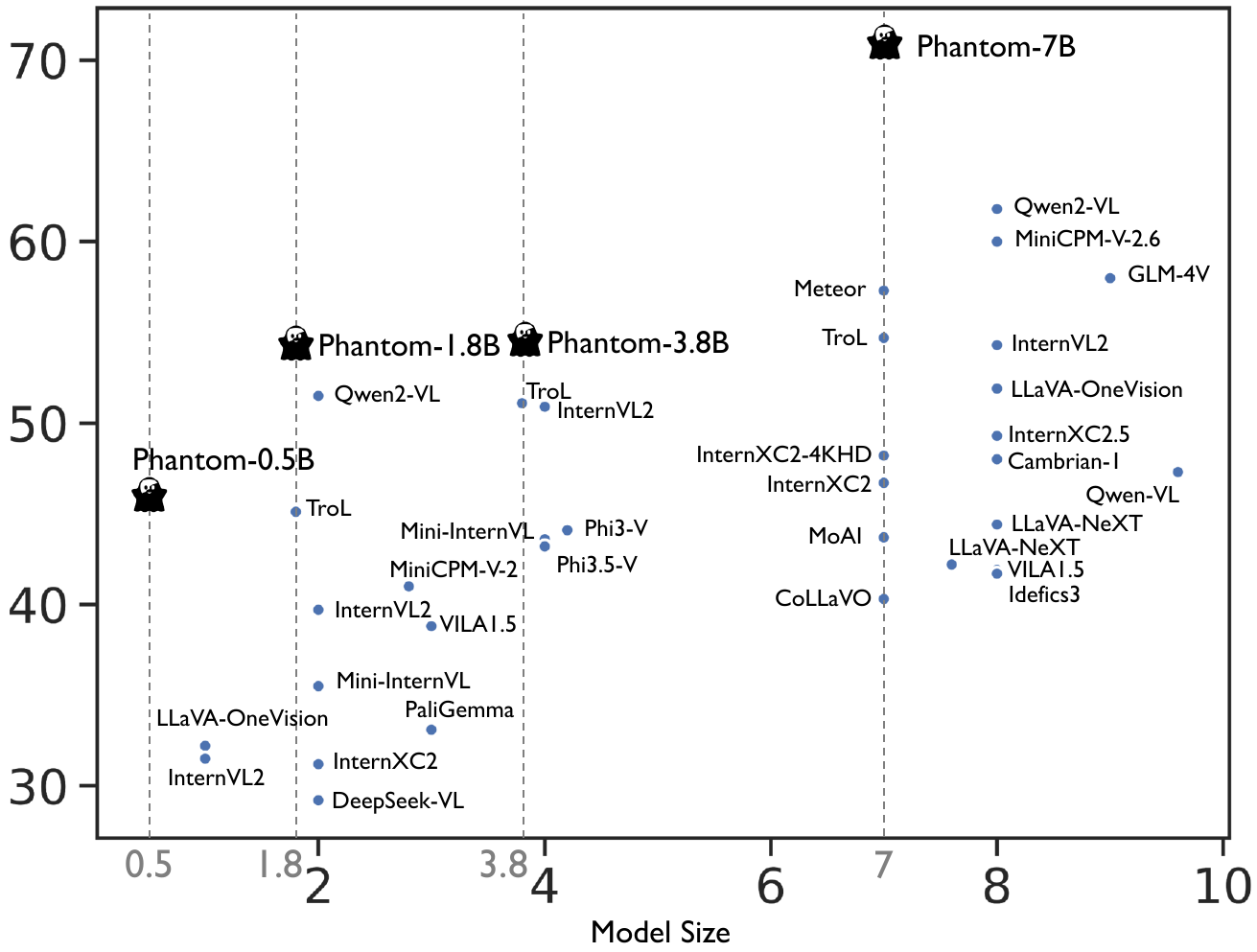}
    \vspace*{-6mm}
    \caption*{(b) 0.5B$\sim$10B}
    \end{minipage}
    }
    \vspace{-3mm}
    \caption{Evaluating MM-Vet~\citep{yu2023mm} for efficient LLVM family, \aphantom Phantom, across four model sizes (0.5B, 1.8B, 3.8B, and 7B), compared with various model size LLVMs: (a) 7B$\sim$80B and unknown model size for closed-source LLVMs (b) 0.5B$\sim$10B model sizes.}
    \label{fig:2}
    \vspace{-6mm}
\end{figure}

%% file: figures/figure3.tex
\begin{figure}[t!]
\vspace{-10mm}
    \centering
    {
    \begin{minipage}[t]{0.45\linewidth}
    \includegraphics[width=\textwidth]{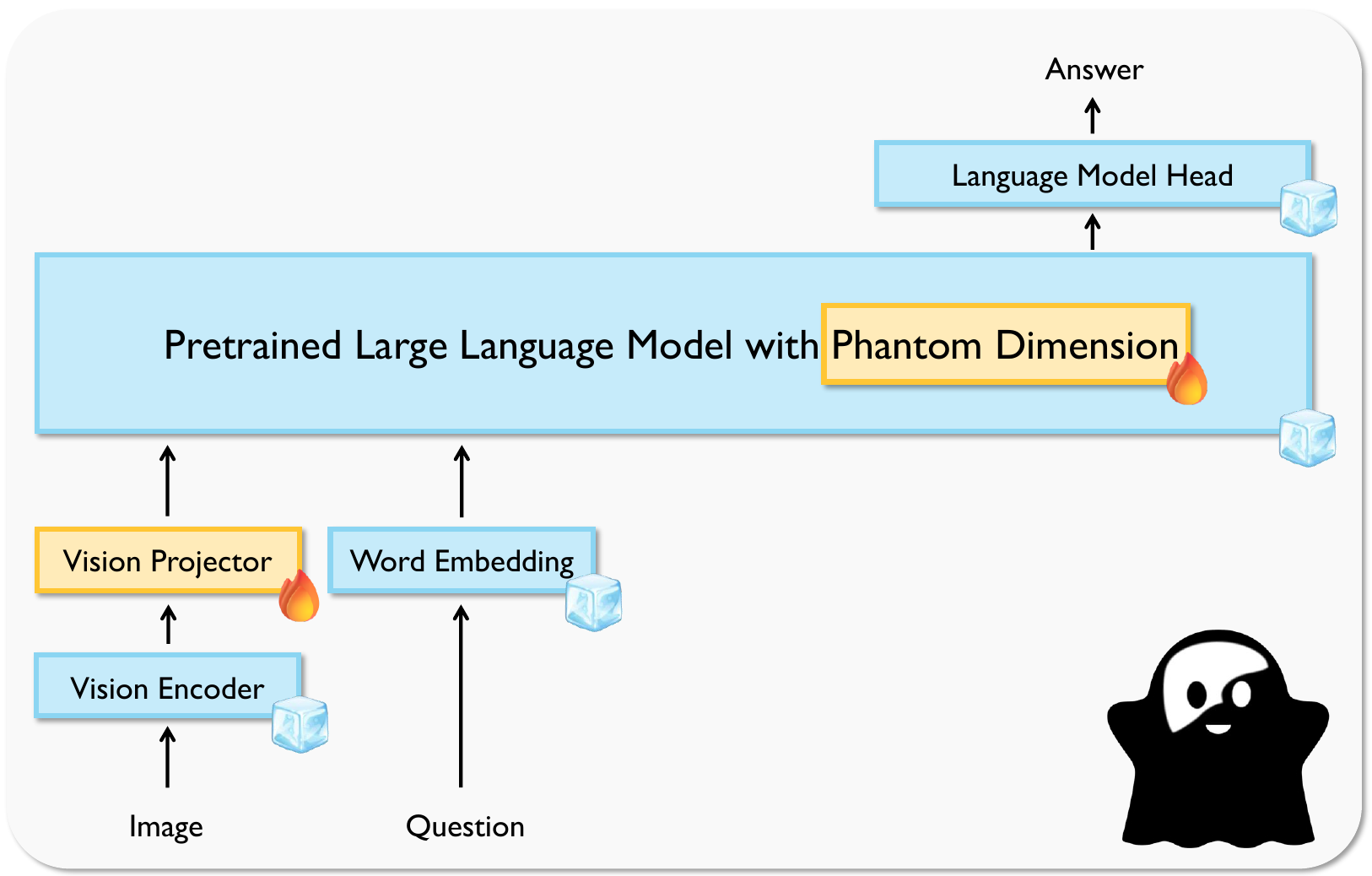}
    \caption*{(a) Model Architecture}
    \end{minipage}
    \begin{minipage}[t]{0.54\linewidth}
    \includegraphics[width=\textwidth]{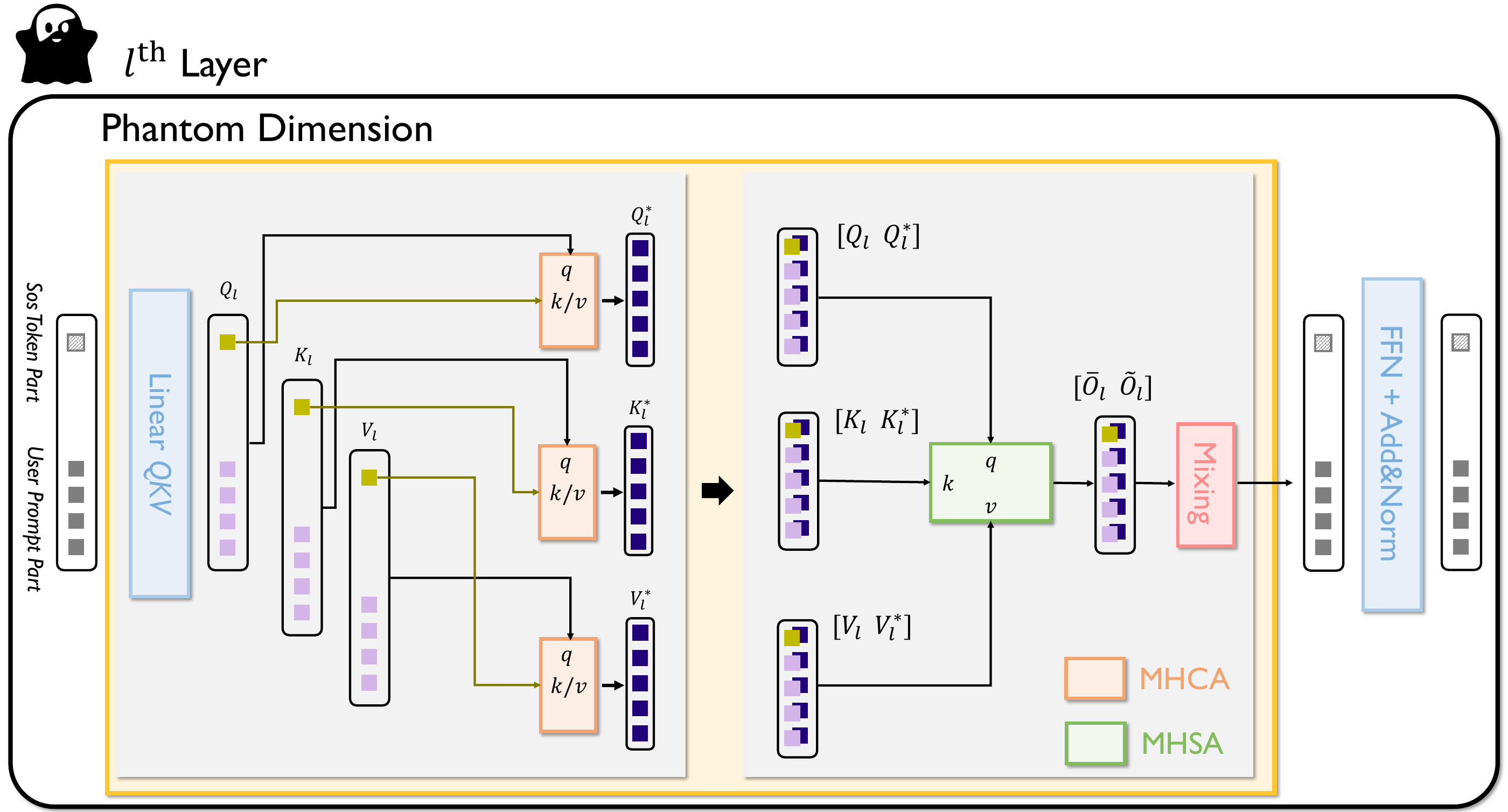}
    \caption*{(b) Phantom Dimension}
    \end{minipage}
    }
    \vspace{-3mm}
    \caption{(a) Overview of model architecture and the detail of first training step with Phantom Dimension and Phantom Optimization. In second training step, we train all of the parameters described in this figure. (b) Illuminating how Phantom Dimension temporarily enlarges the latent hidden dimension in forward propagation at $l$-th layer in \aphantom Phantom, where `Linear QKV', MHSA, and `FFN+Add\&Norm' is generally used module from pretrained LLM. Only MHCA module is added.}
    \label{fig:3}
    \vspace{-3mm}
\end{figure}

%% file: tables/table1.tex
\definecolor{Gray}{gray}{0.93}
\newcommand{\cmark}{\ding{51}}%
\newcommand{\xmark}{\ding{55}}%

\begin{table}[t!]
\vspace{-10mm}
\centering
\caption{Comparison with the current existing standard model size open-source LLVMs, evaluating vision-language performances of \aphantom Phantom on numerous general evaluation benchmarks: SQA$^{\text{I}}$~\citep{lu2022learn}, AI2D~\citep{kembhavi2016diagram}, ChartQA~\citep{masry2022chartqa}, SEED$^{\text{I}}$~\citep{li2023seed}, POPE~\citep{li2023evaluating}, HallB~\citep{liu2023hallusionbench}, MME~\citep{fu2023mme}, MathVista~\citep{lu2023mathvista}, MMB~\citep{liu2023mmbench}, MMB$^{\text{CN}}$~\citep{liu2023mmbench}, MM-Vet~\citep{yu2023mm}, and LLaVA$^{\text{W}}$~\citep{liu2023visual}. \textbf{Bold} and \underline{Underline} represent the top and the second, each.}
\label{tab:1}
\vspace{-3mm}
\resizebox{\linewidth}{!}{
\renewcommand{\tabcolsep}{0.5mm}
\begin{tabular}{lcccccccccccc}
\toprule
LLVMs     & SQA$^{\text{I}}$ & AI2D & ChartQA & SEED$^{\text{I}}$ & POPE & HallB & MME & MathVista & MMB & MMB$^{\text{CN}}$ & MM-Vet & LLaVA$^{\text{W}}$  \\
\midrule
ShareGPT4V-7B~\citep{chen2023sharegpt4v}           & 68.4 & -    & -    & 69.7 & -    & 49.8 & 1944 & 25.8 & 68.8 & 62.2 & 37.6 & -    \\
InternLM-XC-7B~\citep{zhang2023internlm}           & -    & -    & -    & 66.1 & -    & 57.0 & 1919 & 29.5 & 74.4 & 72.4 & 35.2 & -    \\
Monkey-10B~\citep{li2023monkey}                    & 69.4 & -    & -    & 68.9 & -    & 58.4 & 1924 & 34.8 & 72.4 & 67.5 & 33.0 & -    \\
VILA-7B~\citep{lin2023vila}                        & 68.2 & -    & -    & 61.1 & 85.5 & -    & -    & -    & 68.9 & 61.7 & 34.9 & -    \\
VILA-13B~\citep{lin2023vila}                       & 73.7 & -    & -    & 62.8 & 84.2 & -    & -    & -    & 70.3 & 64.3 & 38.8 & -    \\
SPHINX-7B~\citep{lin2023sphinx}                    & 70.6 & -    & -    & 71.6 & 86.9 & -    & 1797 & 27.8 & 65.9 & 57.9 & 40.2 & -    \\
SPHINX-MoE-7B$\times$8~\citep{gao2024sphinx}       & 70.6 & -    & -    & 73.0 & \textbf{89.6} & -    & 1852 & 42.7 & 71.3 & -    & 40.9 & -    \\
SPHINX-Plus-13B~\citep{gao2024sphinx}              & 70.6 & -    & -    & 74.8 & \underline{89.1} & 52.1 & 1741 & 36.8 & 71.0 & -    & 47.9 & -    \\
LLaVA-NeXT-7B~\citep{liu2024llavanext}             & 70.1 & -    & -    & 70.2 & 86.5 & -    & 1851 & 34.6 & 69.6 & 63.3 & 43.9 & 72.3 \\
LLaVA-NeXT-8B~\citep{liu2024llavanext}             & -    & 71.6 & 69.5 & -    & -    & -    & 1972 & 37.5 & 72.1 & -    & -    & 80.1 \\
LLaVA-NeXT-13B~\citep{liu2024llavanext}            & 73.6 & 70.0 & 62.2 & 72.2 & 86.7 & -    & 1892 & 35.1 & 70.0 & 68.5 & 47.3 & 72.3 \\
MM1-7B~\citep{mckinzie2024mm1}                     & 72.6 & -    & -    & 69.9 & 86.6 & -    & 1858 & 35.9 & 72.3 & -    & 42.1 & -    \\
MM1-MoE-7B$\times$32~\citep{mckinzie2024mm1}       & 74.4 & -    & -    & 70.9 & 87.8 & -    & 1992 & 40.9 & 72.7 & -    & 45.2 & -    \\
MiniGemini-HD-7B~\citep{li2024mini}                & -    & -    & -    & -    & -    & -    & 1865 & 32.2 & 65.8 & -    & 41.3 & -    \\
MiniGemini-HD-13B~\citep{li2024mini}               & -    & -    & -    & -    & -    & -    & 1917 & 37.0 & 68.6 & -    & 50.5 & -    \\
Cambrian-1-8B~\citep{tong2024cambrian}
            & 80.4 % SQA
            & 73.0 % AI2D
            & 73.3 % ChartQA
            & 74.7 % SEED
            & - % POPE
            & - % HallB    
            & - % MME
            & 49.0 % MathVista
            & 75.9 %MMB
            & - %MMB CN
            & - %MM-Vet
            & - \\ %LLaVA-W   
Cambrian-1-13B~\citep{tong2024cambrian}  
            & 79.3 % SQA
            & 73.6 % AI2D
            & 73.8 % ChartQA
            & 74.4 % SEED
            & - % POPE
            & - % HallB    
            & - % MME
            & 48.0 % MathVista
            & 75.7 %MMB
            & - %MMB CN
            & - %MM-Vet
            & - \\ %LLaVA-W 
Eagle-8B~\citep{shi2024eagle}  
            & 84.3 % SQA
            & 76.1 % AI2D
            & 80.1 % ChartQA
            & 76.3 % SEED
            & - % POPE
            & - % HallB    
            & - % MME
            & 52.7 % MathVista
            & 75.9 %MMB
            & - %MMB CN
            & - %MM-Vet
            & - \\ %LLaVA-W 
Eagle-13B~\citep{shi2024eagle}   
            & 82.0 % SQA
            & 74.0 % AI2D
            & 77.6 % ChartQA
            & 74.8 % SEED
            & - % POPE
            & - % HallB    
            & - % MME
            & 54.4 % MathVista
            & 75.7 %MMB
            & - %MMB CN
            & - %MM-Vet
            & - \\ %LLaVA-W 
VILA1.5-8B~\citep{lin2023vila}  
            & 82.0 
            & -    
            & -
            & 73.8 
            & 85.6
            & -    
            & -
            & -
            & 75.3 
            & 69.9  
            & 43.2 
            & 71.9 \\
VILA1.5-13B~\citep{lin2023vila}  
            & 80.1 
            & -    
            & -
            & 72.6 
            & 86.3
            & -    
            & -
            & -
            & 74.9 
            & 66.3 
            & 44.3 
            & 80.8 \\
VILA$^2$-8B~\citep{fang2024vila}  
            & 87.6
            & -    
            & -
            & 66.1 
            & 86.7
            & -    
            & -
            & -
            & 76.6 
            & 71.7  
            & 50.0 
            & 86.6 \\
MiniCPM-V-2.5-8B~\citep{yao2024minicpm}
            & - % SQA
            & - % AI2D
            & - % ChartQA
            & - % SEED
            & - % POPE
            & - % HallB    
            & 2025 % MME
            & 54.3  % MathVista
            & 77.2 %MMB
            & 74.2 %MMB CN
            & - %MM-Vet
            & 86.7 \\ %LLaVA-W 
CogVLM2-8B~\citep{hong2024cogvlm2}
            & - % SQA
            & 73.4 % AI2D
            & 81.0 % % ChartQA
            & - % SEED
            & - % POPE
            & - % HallB    
            & 1870 % MME
            & -  % MathVista
            & 80.5 %MMB
            & - %MMB CN
            & \underline{60.4} %MM-Vet
            & - \\ %LLaVA-W 
TroL-7B~\citep{lee2024trol}
            & 92.8 % SQA
            & 78.5 % AI2D
            & 71.2 % ChartQA
            & \underline{75.3} % SEED
            & 87.8 % POPE
            & \underline{65.3} % HallB    
            & \textbf{2308} % MME
            & 51.8  % MathVista
            & \underline{83.5} %MMB
            & \underline{81.2} %MMB CN
            & 54.7 %MM-Vet
            & \textbf{92.8} \\ %LLaVA-W 
LLaVA-OneVision-8B~\citep{li2024llava}
            & \textbf{96.0} % SQA
            & \textbf{81.4} % AI2D
            & 80.0% % ChartQA
            & \textbf{75.4} % SEED
            & - % POPE
            & - % HallB    
            & 1998 % MME
            & \underline{63.2}  % MathVista
            & 80.8 %MMB
            & - %MMB CN
            & 57.5 %MM-Vet
            & \underline{90.7} \\ %LLaVA-W 
\midrule
\rowcolor{Gray}
\aphantom
Phantom-7B  
            & \underline{95.6}
            & \underline{79.5}    
            & \textbf{87.8}
            & \underline{75.3}
            & 87.7
            & \textbf{65.4}    
            & \underline{2126}
            & \textbf{70.9}
            & \textbf{84.8} 
            & \textbf{84.7}  
            & \textbf{70.8} 
            & 84.9    \\
\bottomrule 
\end{tabular}
}
\vspace{-7mm}
\end{table}

%% file: tables/table2.tex
\begin{table}[t!]
\vspace{-10mm}
\centering
\caption{Comparison with the current existing smaller open-source LLVMs across 0.5B$\sim$4B model sizes, evaluating vision-language performances of \aphantom Phantom on numerous evaluation benchmarks equally used in Table~\ref{tab:1}.}
\label{tab:2}
\vspace{-3mm}
\resizebox{\linewidth}{!}{
\renewcommand{\tabcolsep}{0.5mm}
\begin{tabular}{lcccccccccccc}
\toprule
LLVMs     & SQA$^{\text{I}}$ & AI2D & ChartQA & SEED$^{\text{I}}$ & POPE & HallB & MME & MathVista & MMB & MMB$^{\text{CN}}$ & MM-Vet & LLaVA$^{\text{W}}$  \\
\midrule
%                                                  & SQA  & AI2D & ChQA & SEED & POPE & Hall & MME  & MatV & MMB  & MMBC & MMVet& llava
% UIO-2-XL-3.2B~\citep{lu2023unifiedio}              & 87.4 & -    & -    & 60.2 & 87.2 & -    & -    & -    & 68.1 & -    & -    & -       \\
% Gemini Nano-2-3.2B~\citep{team2023gemini}          & -    & -    & -    & -    & -    & -    & -    & 30.6 & -    & -    & -    & -       \\
MobileVLM-3B~\citep{chu2023mobilevlm}              & 61.2 & -    & -    & -    & 84.9 & -    & -    & -    & 59.6 & -    & -    & -       \\
MobileVLM-V2-3B~\citep{chu2024mobilevlm}           & 70.0 & -    & -    & -    & 84.7 & -    & -    & -    & 63.2 & -    & -    & -       \\
MoE-LLaVA-2.7B$\times$4~\citep{lin2024moe}         & 70.3 & -    & -    & -    & 85.7 & -    & -    & -    & 68.0 & -    & 35.9 & -       \\
LLaVA-Phi-2.7B~\citep{zhu2024llava}                & 68.4 & -    & -    & -    & 85.0 & -    & -    & -    & 59.8 & -    & 28.9 & -       \\
Imp-v1-3B~\citep{shao2024imp}                      & 70.0 & -    & -    & -    & \textbf{88.0} & -    & -    & -    & 66.5 & -    & 33.1 & -       \\
TinyLLaVA-3.1B~\citep{zhou2024tinyllava}           & 69.1 & -    & -    & -    & 86.4 & -    & -    & -    & 66.9 & -    & 32.0 & -       \\
TinyLLaVA-Sig-Phi-3.1B~\citep{zhou2024tinyllava}   & 69.1 & -    & -    & -    & 86.4 & -    & -    & -    & 66.9 & -    & 32.0 & -       \\
Bunny-3B~\citep{he2024efficient}                   & 70.9 & 38.2 & -    & 62.5 & 86.8 & -    & 1778 & -    & 68.6 & -    & -    & -       \\
MiniCPM-2.4B~\citep{hu2024minicpm}                 & -    & 56.3 & -    & -    & -    & -    & 1650 & 28.9 & 64.1 & 62.6 & 31.1 & -       \\
MiniCPM-V2-2.8B~\citep{hu2024minicpm}              & -    & 62.9 & -    & -    & -    & -    & 1809 & 38.7 & 69.1 & 66.5 & 41.0 & -       \\
MM1-3B~\citep{mckinzie2024mm1}                     & 69.4 & -    & -    & 68.8 & 87.4 & -    & 1762 & 32.0 & 67.8 & -    & 43.7 & -       \\
MM1-MoE-3B$\times$64~\citep{mckinzie2024mm1}       & 76.1 & -    & -    & 69.4 & \underline{87.6} & -    & 1773 & 32.6 & 70.8 & -    & 42.2 & -       \\
ALLaVA-3B~\citep{chen2024allava}                   & -    & -    & -    & 65.2 & -    & -    & 1623 & -    & 64.0 & -    & 32.2 & -       \\
ALLaVA-3B-Longer~\citep{chen2024allava}            & -    & -    & -    & 65.6 & -    & -    & 1564 & -    & 64.6 & -    & 35.5 & -       \\
VILA1.5-3B~\citep{chen2024allava}                  & 69.6 & -    & -    & 66.4 & 85.3 & -    & -    & -    & 62.8 & 52.2 & 38.6 & \textbf{76.7}    \\
TroL-3.8B~\citep{lee2024trol}
            & \underline{90.8}  % SQA
            & \textbf{73.6}  % AI2D
            & \underline{73.8} % ChartQA
            & \underline{70.5} % SEED
            & 86.5 % POPE
            & \textbf{62.2} % HallB    
            & \underline{1980} % MME
            & \underline{55.1}  % MathVista
            & \underline{79.2} %MMB
            & \textbf{77.1}  %MMB CN
            & \underline{51.1} %MM-Vet
            & \underline{76.6} \\ %LLaVA-W 
\midrule
\rowcolor{Gray}
\aphantom
Phantom-3.8B 
            & \textbf{94.2} % SQA
            & \underline{71.7} % AI2D  
            & \textbf{87.3} % ChQA
            & \textbf{72.8} % SEED
            & 87.1 % POPE
            & \underline{60.8} % Hall  
            & \textbf{2046} % MME
            & \textbf{60.6} % MatV
            & \textbf{80.4} % MMB
            & \textbf{77.1} % MMBC
            & \textbf{54.4} % MMVet
            & 76.2 \\ % llava
\midrule
% UIO-2-L-1.1B~\citep{lu2023unifiedio}               & 78.6 & -    & -    & 51.1 & 77.8 & -    & -    & -    & 62.1 & -    & -    & -        \\
DeepSeek-VL-1.3B~\citep{lu2024deepseek}            & -    & -    & -    & 66.7 & 87.6 & -    & -    & 31.1 & 64.6 & 62.9 & 34.8 & -        \\
MobileVLM-1.7B~\citep{chu2023mobilevlm}            & 57.3 & -    & -    & -    & 84.5 & -    & -    & -    & 53.2 & -    & -    & -        \\
MobileVLM-V2-1.7B~\citep{chu2024mobilevlm}         & 66.7 & -    & -    & -    & 84.3 & -    & -    & -    & 57.7 & -    & -    & -        \\
MoE-LLaVA-1.8B$\times$4~\citep{lin2024moe}         & 63.1 & -    & -    & -    & 87.0 & -    & -    & -    & 59.7 & -    & 25.3 & -       \\
Mini-Gemini-2B~\citep{li2024mini}                  & -    & -    & -    & -    & -    & -    & 1653 & 29.4 & 59.8 & -    & -    & -       \\
TroL-1.8B~\citep{lee2024trol}
            & \underline{87.5} % SQA
            & \textbf{68.9} % AI2D
            & \underline{64.0} % ChartQA
            & \textbf{69.0}  % SEED
            & \underline{88.6} % POPE
            & \underline{60.1}  % HallB    
            & \textbf{2038} % MME
            & \underline{45.4}   % MathVista
            & \underline{76.1} %MMB
            & \underline{74.1}  %MMB CN
            & \underline{45.1}  %MM-Vet
            & \textbf{69.7} \\ %LLaVA-W 
\midrule
\rowcolor{Gray}
\aphantom
Phantom-1.8B 
            & \textbf{91.9} % SQA
            & \underline{62.3} % AI2D  
            & \textbf{87.0} % ChQA
            & \underline{68.6} % SEED
            & \textbf{89.6} % POPE
            & \textbf{62.2} % Hall  
            & \underline{1885} % MME
            & \textbf{60.9} % MatV
            & \textbf{76.6} % MMB
            & \textbf{75.1} % MMBC
            & \textbf{54.1} % MMVet
            & \underline{68.6} \\ % llava
\midrule
LLaVA-OneVision-0.5B~\citep{li2024llava}         & 67.2 & \textbf{57.1}    & 61.4    & \textbf{65.5}    & -   & -    & 1478    & 34.8    & 52.1 & -    & 29.1    & \textbf{74.2}        \\
\midrule
\rowcolor{Gray}
\aphantom
Phantom-0.5B 
            & \textbf{83.2} % SQA
            & 54.1 % AI2D  
            & \textbf{78.0} % ChQA
            & 60.6 % SEED
            & \textbf{86.0} % POPE
            & \textbf{54.6} % Hall  
            & \textbf{1743} % MME
            & \textbf{51.7} % MatV
            & \textbf{72.7} % MMB
            & \textbf{70.1} % MMBC
            & \textbf{45.7} % MMVet
            & 69.6    \\ % llava
\bottomrule

\end{tabular}
}
\vspace{-7mm}
\end{table}

%% file: tables/table3.tex
\begin{table}[t!]
\vspace{-10mm}
\centering
\caption{Detailed comparison for challenging evaluation benchmarks. Sub-benchmark category names in (c), (d), and (g) are represented in Appendix~\ref{app:B}. For (f), LLaVA-Wilder~\citep{zhang2024lmms} is a more advanced challenging evaluation benchmark over LLaVA$^{\text{W}}$~\citep{liu2023visual}.}
\label{tab:3}
\vspace{-3mm}
\begin{minipage}[t]{\linewidth}

\begin{minipage}[t]{0.99\linewidth}
\input{tables/table3a}
\end{minipage}

\begin{minipage}[t]{0.99\linewidth}
\input{tables/table3b}
\end{minipage}

{\begin{minipage}[t]{0.49\linewidth}
\input{tables/table3c}
\end{minipage}
\begin{minipage}[t]{0.49\linewidth}
\input{tables/table3d}
\end{minipage}
}

{\begin{minipage}[t]{0.65\linewidth}
\input{tables/table3e}
\end{minipage}
\begin{minipage}[t]{0.34\linewidth}
\input{tables/table3f}
\end{minipage}
}

{\begin{minipage}[t]{0.58\linewidth}
\input{tables/table3g}
\end{minipage}
\begin{minipage}[t]{0.42\linewidth}
\input{tables/table3h}
\end{minipage}
}

\end{minipage}
\vspace{-7mm}
\end{table}

%% file: tables/table3a.tex
\newcolumntype{g}{>{\columncolor{Gray}}c}
\caption*{(a) Comparison with LLVMs using additional modules and projector: OmniFusion~\cite{goncharova2024omnifusion}, DeepSeek-VL~\citep{lu2024deepseek}, MoVA~\citep{kar2024brave}, Eagle~\citep{shi2024eagle}, CoLLaVO~\citep{lee2024collavo}, MoAI~\citep{lee2024moai}, and Meteor~\citep{lee2024meteor}}
\vspace{-3mm}
\resizebox{\linewidth}{!}{
\renewcommand{\tabcolsep}{1mm}
\begin{tabular}{lcccccccg}
\toprule
Benchmarks                              & OmniFusion-7B & DeepSeek-VL-7B & MoVA-7B & Eagle-8B & CoLLaVO-7B & MoAI-7B & Meteor-7B & \textbf{Phantom-7B} \\
\midrule
SQA$^{\text{I}}$~\citep{lu2022learn}    & 69.7          & 57.7           & 74.4    & 84.3     & 80.7       & 83.5    & \underline{87.5}      & \textbf{95.6}\\
\cdashline{1-9}\noalign{\vskip 0.5ex}
MMB~\citep{liu2023mmbench}              & 69.0          & 73.2           & 81.3    & 75.9     & \underline{83.0}       & 79.3    & 82.9      & \textbf{84.8}\\
\cdashline{1-9}\noalign{\vskip 0.5ex}
MM-Vet~\citep{yu2023mm}                 & 39.4          & 41.5           & -       & -        & 40.3       & 43.7    & \underline{57.3}      & \textbf{70.8}\\
\cdashline{1-9}\noalign{\vskip 0.5ex}
MathVista~\citep{lu2023mathvista}       & -             & -              & 44.3    & 52.7     & \underline{57.6}       & 56.2    & 53.4      & \textbf{70.9}\\
\cdashline{1-9}\noalign{\vskip 0.5ex}
MMStar~\citep{chen2024we}               & -             &  -             & -       & -        & 42.1       & \underline{48.7}    & 45.5      & \textbf{57.7}\\
\bottomrule
\end{tabular}
}

%% file: tables/table3b.tex
\newcolumntype{g}{>{\columncolor{Gray}}c}
\caption*{(b) Comparison on challenging evaluation benchmarks with more recently released open-source LLVMs: Cambrian-1~\citep{tong2024cambrian}, LLaVA-OneVision(OV)~\citep{li2024llava}, MiniCPM-V-2.6~\citep{yao2024minicpm}, InternVL2~\citep{chen2024far}, and Qwen2-VL~\citep{wang2024qwen2vlenhancingvisionlanguagemodels}, which are trained on larger datasets and with greater computational resources, alongside GPT-4V.}
\vspace{-3mm}
\resizebox{\linewidth}{!}{
\renewcommand{\tabcolsep}{2mm}
\begin{tabular}{lccccccg}
\toprule
Benchmarks                             & Cambrian-1-8B   & LLaVA-OV-8B       & MiniCPM-V-2.6-7B     & InternVL2-8B  & Qwen2-VL-7B  & GPT-4V & \textbf{Phantom-7B} \\
\midrule
CV-Bench~\citep{tong2024cambrian}      & \underline{72.2}            & -                 & -                    & -             & -            & 69.1   & \textbf{74.9}\\
\cdashline{1-8}\noalign{\vskip 0.5ex}
BLINK~\citep{fu2024blink}              & 44.9            & 48.2              & -                    & 50.9          & -            & \underline{58.3}   & \textbf{58.9}\\
\cdashline{1-8}\noalign{\vskip 0.5ex}
MM-Vet~\citep{yu2023mm}                & 51.7            & 57.5              & 60.0                 & 60.0          & 62.0         & \underline{63.6}   & \textbf{70.8}\\
\cdashline{1-8}\noalign{\vskip 0.5ex}
ChartQA~\citep{masry2022chartqa}       & 73.3            & 80.0              & -                    & \underline{83.3}          & 83.0         & 78.5   & \textbf{87.8}\\
\cdashline{1-8}\noalign{\vskip 0.5ex}
MathVista~\citep{lu2023mathvista}      & 49.0            & -                 & 60.6                 & 58.3          & 58.2         & \underline{69.1}   & \textbf{70.9}\\
\bottomrule
\end{tabular}
}

%% file: tables/table3c.tex
\caption*{(c) MMStar~\citep{chen2024we}}
\vspace{-3mm}
\resizebox{\linewidth}{!}{
\renewcommand{\tabcolsep}{1mm}
\begin{tabular}{lccccccc}
\toprule
LLVMs            & CP                       & FP                       & IR                       & LR                       & ST                       & MA                       & Avg                      \\
\midrule
Yi-VL-34B~\citep{young2024yi}        & 53.2 & 31.2 & 52.0 & 32.4 & 12.4 & 35.2 & 36.1 \\
\cdashline{1-8}\noalign{\vskip 0.5ex}
CogVLM-Chat-17B~\citep{wang2023cogvlm}  & 66.8 & 36.8 & 49.2 & 31.2 & 23.6 & 11.6 & 36.5 \\
\cdashline{1-8}\noalign{\vskip 0.5ex}
SPHINX-MoE-7B$\times$8~\citep{gao2024sphinx} & 58.4                     & 40.8                     & 47.6                     & 35.2                     & 19.2                     & 32.0                     & 38.9                     \\
\cdashline{1-8}\noalign{\vskip 0.5ex}
InternVL1.2-40B~\citep{chen2023internvl}  & 67.6                     & 43.2                     & 61.2                     & 47.2                     & 24.0                     & 19.2                     & 43.7                     \\
\cdashline{1-8}\noalign{\vskip 0.5ex}
LLaVA-NeXT-34B~\citep{liu2024llavanext}   & 66.4                     & \underline{52.0}            & 62.4                     & 46.0                     & 32.4                     & \underline{53.6}            & 52.1                     \\
\cdashline{1-8}\noalign{\vskip 0.5ex}
InternXC2-7B~\citep{dong2024internlm}   & \underline{70.8}                     & 48.8            & \underline{65.2}                     & \underline{56.4}                     & \underline{42.0}                     & 49.2            & 55.4                     \\
\cdashline{1-8}\noalign{\vskip 0.5ex}
GPT-4V~\citep{gptsyscard}   & \textbf{76.6}                     & 51.4            & \textbf{66.6}                     & 55.8                     & \textbf{42.6}                     & 49.8            & \underline{57.1}                     \\

\midrule
% \cdashline{1-8}\noalign{\vskip 0.5ex}
\rowcolor{Gray}
\textbf{Phantom-7B}     
               & 66.0
               & \textbf{52.8}
               & 60.0
               & \textbf{60.8}
               & 38.4
               & \textbf{68.4}
               & \textbf{57.7}        \\
\bottomrule
\end{tabular}
}

%% file: tables/table3d.tex
\caption*{(d) MathVerse~\citep{zhang2024mathverse}}
\vspace{-3mm}
\resizebox{\linewidth}{!}{
\renewcommand{\tabcolsep}{1mm}
\begin{tabular}{lccccccc}
\toprule
LLVMs                                          & TD   & TL   & TO   & VI   & VD   & VO   & Avg  \\
\midrule
G-LLaVA-7B~\citep{gao2023g}                    & 20.9 & 20.7 & 21.1 & 17.2 & 16.4 & 9.4  & 16.6 \\
\cdashline{1-8}\noalign{\vskip 0.5ex}
LLaVA-NeXT-13B~\citep{liu2024llavanext}        & 12.8 & 12.0 & 9.9  & 10.7 & 9.7  & 6.3  & 10.3 \\
\cdashline{1-8}\noalign{\vskip 0.5ex}
ShareGPT4V-13B~\citep{chen2023sharegpt4v}      & 16.2 & 16.2 & 6.6  & 15.5 & 13.8 & 3.7  & 13.1 \\
\cdashline{1-8}\noalign{\vskip 0.5ex}
SPHINX-MoE-7B$\times$8~\citep{gao2024sphinx}   & 26.2 & 17.4 & 26.7 & 16.7 & 12.5 & 11.1 & 16.8 \\
\cdashline{1-8}\noalign{\vskip 0.5ex}
InternXC2-7B~\citep{dong2024internlm}          & 22.3 & 17.0 & 16.5 & 15.7 & 16.4 & 11.0 & 16.5 \\
\cdashline{1-8}\noalign{\vskip 0.5ex}
LLaVA-NeXT-34B~\citep{liu2024llavanext}        & 33.8 & 25.5 & 21.3 & 23.5 & 20.3 & 15.7 & 23.8 \\
\cdashline{1-8}\noalign{\vskip 0.5ex}
GPT-4V~\citep{gptsyscard}                      & \textbf{54.7} & \underline{41.4} & \textbf{48.7} & \underline{34.9} & \underline{34.4} & \underline{31.6} & \underline{39.4} \\
\midrule
% \cdashline{1-8}\noalign{\vskip 0.5ex}
\rowcolor{Gray}
\textbf{Phantom-7B}                      & \underline{47.3}
                                         & \textbf{45.2}
                                         & \underline{45.3}
                                         & \textbf{42.7}
                                         & \textbf{41.7}
                                         & \textbf{43.7}
                                         & \textbf{41.0} \\
\bottomrule
\end{tabular}
}

%% file: tables/table3e.tex
\caption*{(e) MM-Vet-v2~\citep{yu2024mm}}
\vspace{-3mm}
\resizebox{\linewidth}{!}{
\renewcommand{\tabcolsep}{1.5mm}
\begin{tabular}{lcccccccc}
\toprule
LLVMs                                           & Rec    & Gen    & OCR    & Spat     & Know    & Seq     & Math     & Avg                      \\
\midrule
LLaVA-NeXT-34B~\citep{liu2024llavanext}         & 49.3   & 48.9   & 53.2   & 48.3     & 49.6    & 18.5    & 37.3     & 50.9                        \\
\cdashline{1-9}\noalign{\vskip 0.5ex}
InternVL-Chat-V1-5~\citep{chen2024far}          & 52.0   & 48.9   & 51.7   & 49.3     & 47.9    & 37.6    & 17.6     & 51.5                        \\
\cdashline{1-9}\noalign{\vskip 0.5ex}
Claude3 Opus~\citep{claude3series2024}          & 53.5   & \textbf{57.6}   & 60.5   & 50.0     & 51.0    & \textbf{46.1}    & 45.6     & 55.8                         \\
\cdashline{1-9}\noalign{\vskip 0.5ex}
Qwen-VL-Max~\citep{bai2023qwen}                 & 51.7   & 51.1   & 60.2   & 49.0     & \textbf{52.2}    & 27.3    & \underline{58.3}     & 55.8                         \\
\cdashline{1-9}\noalign{\vskip 0.5ex}
Gemini-Pro~\citep{team2023gemini}               & \underline{54.3}   & 50.8   & \underline{61.9}   & \underline{55.8}     &  50.7   & \underline{45.4}    & 46.3     &  \underline{57.2}                        \\
\midrule
% \cdashline{1-8}\noalign{\vskip 0.5ex}
\rowcolor{Gray}
\textbf{Phantom-7B}     & \textbf{56.1}
               & \underline{53.9}
               & \textbf{67.4}
               & \textbf{57.7}
               & \underline{51.9}
               & 37.3
               & \textbf{68.5}
               & \textbf{60.6} \\
\bottomrule
\end{tabular}
}

%% file: tables/table3f.tex
\caption*{(f) LLaVA-Wilder}
\vspace{-3mm}
\resizebox{\linewidth}{!}{
\renewcommand{\tabcolsep}{3mm}
\begin{tabular}{lc}
\toprule
LLVMs                                   &   Accuracy\\
\midrule
LLaVA-NeXT-8B~\citep{liu2024llavanext}   &  62.5 \\
\cdashline{1-2}\noalign{\vskip 0.5ex}
LLaVA-NeXT-72B~\citep{liu2024llavanext}  & 71.2 \\
\cdashline{1-2}\noalign{\vskip 0.5ex}
LLaVA-NeXT-110B~\citep{liu2024llavanext} & 70.5 \\
\cdashline{1-2}\noalign{\vskip 0.5ex}
LLaVA-OV-7B~\citep{li2024llava}          & 67.8 \\
\cdashline{1-2}\noalign{\vskip 0.5ex}
LLaVA-OV-72B~\citep{li2024llava}         & \underline{72.0} \\
\cdashline{1-2}\noalign{\vskip 0.5ex}
GPT-4V~\citep{gptsyscard}                & 71.5 \\
\midrule
\rowcolor{Gray}
\textbf{Phantom-7B}     & \textbf{83.7} \\
\bottomrule
\end{tabular}
}

%% file: tables/table3g.tex
\caption*{(g) VisualWebBench~\cite{liu2024visualwebbench}.}
\vspace{-3mm}
\resizebox{\linewidth}{!}{
\renewcommand{\tabcolsep}{1mm}
\begin{tabular}{lcccccccc}
\toprule
\multirow{2}{*}{LLVMs} & \multicolumn{3}{c}{Website} & \multicolumn{2}{c}{Element} & \multicolumn{2}{c}{Action} & \multirow{2}{*}{Average} \\
\cmidrule(lr){2-4}\cmidrule(lr){5-6}\cmidrule(lr){7-8}
                                           & Cap       & QA    & OCR     & OCR         & Grd           & Pred           & Grd    &                          \\
\midrule
LLaVA-NeXT-7B~\citep{liu2024llavanext}      & 27.0     & 39.8  & 57.3    & 54.8        & 31.7          & 30.6           & 10.7      & 36.0                     \\
\cdashline{1-9}\noalign{\vskip 0.5ex}
LLaVA-NeXT-13B~\citep{liu2024llavanext}     & 26.5     & 44.5  & 52.8    & 56.1        & 31.7          & 48.4           & 15.5      & 39.4                     \\
\cdashline{1-9}\noalign{\vskip 0.5ex}
LLaVA-NeXT-34B~\citep{liu2024llavanext}     & 24.3     & 48.2  & 67.1    & 71.9        & 43.1          & 74.0           & 25.2      & 50.5                     \\
\cdashline{1-9}\noalign{\vskip 0.5ex}
Gemini-Pro~\citep{team2023gemini}           & 25.0     & 55.5  & \underline{75.1}    & 65.4        & 44.3          & 26.7           & 43.7      & 48.0                     \\
\cdashline{1-9}\noalign{\vskip 0.5ex}
Claude3 Sonnet~\citep{claude3series2024}    & 28.9     & \textbf{81.8}  & 70.3    & \textbf{89.2}        & \underline{68.8}          & 63.4           & 58.3      & \underline{65.8}                     \\
\cdashline{1-9}\noalign{\vskip 0.5ex}
Claude3 Opus~\citep{claude3series2024}      & 26.7     & \underline{75.4}  & 63.7    & \underline{87.1}        & 57.7          & 60.4           & 38.8      & 58.5                     \\
\cdashline{1-9}\noalign{\vskip 0.5ex}
GPT-4V~\citep{gptsyscard}                   & \textbf{34.5}     & 75.0  & 68.8    & 62.8        & 67.5          & \underline{67.6}           & \textbf{75.7}      & 64.6                     \\
\midrule
% \cdashline{1-8}\noalign{\vskip 0.5ex}
\rowcolor{Gray}
\textbf{Phantom-7B}    & \underline{29.0}    
                       & 70.2 
                       & \textbf{73.8}   
                       & 72.3
                       & \textbf{82.8}
                       & \textbf{78.6}
                       & \underline{66.9}
                       & \textbf{67.7}       \\
\bottomrule
\end{tabular}
}

%% file: tables/table3h.tex
\caption*{(h) SEED-Bench-2-Plus~\citep{li2024seed}}
\vspace{-3mm}
\resizebox{\linewidth}{!}{
\renewcommand{\tabcolsep}{0.8mm}
\begin{tabular}{lcccc}
\toprule
LLVMs                                    & Charts & Maps & Webs & Acc\\
\midrule
LLaVA-NeXT-7B~\citep{liu2024llavanext}   & 36.4 & 34.0 & 39.9 & 36.8 \\
\cdashline{1-5}\noalign{\vskip 0.5ex}
SPHINX2-13B~\citep{gao2024sphinx}        & 41.7 & 41.9 & \underline{60.5} & 48.0 \\
\cdashline{1-5}\noalign{\vskip 0.5ex}
InternXC-7B~\citep{zhang2023internlm}    & 39.9 & 39.0 & 43.0 & 40.6 \\
\cdashline{1-5}\noalign{\vskip 0.5ex}
InternXC2-7B~\citep{dong2024internlm}    & 49.4 & 47.1 & 58.0 & 51.5 \\
\cdashline{1-5}\noalign{\vskip 0.5ex}
SEED-X-13B~\citep{ge2024seed}            & 46.9 & 43.3 & 52.6 & 47.1 \\
\cdashline{1-5}\noalign{\vskip 0.5ex}
Gemini-Pro~\citep{team2023gemini}        & 52.1 & \underline{49.4} & 56.8 & 52.8 \\
\cdashline{1-5}\noalign{\vskip 0.5ex}
Claude3 Opus~\citep{claude3series2024}   & 43.7 & 43.9 & 45.1 & 44.2 \\
\cdashline{1-5}\noalign{\vskip 0.5ex}
GPT-4V~\citep{gptsyscard}                & \underline{54.8} & \underline{49.4} & 57.2 & \underline{53.8} \\
\midrule
\rowcolor{Gray}
\textbf{Phantom-7B}                      & \textbf{62.5} & \textbf{56.4} & \textbf{80.5} & \textbf{65.5} \\
\bottomrule
\end{tabular}
}

%% file: tables/table4.tex
% underline function
\newcommand{\uu}[1]{\underline{#1}}

\begin{table}[t!]
\vspace{-10mm}
\centering
\caption{Identifying the effectiveness of \aphantom Phantom by controlling the three factors: Weighted-Average (WA) operation, Phantom Dimension (PD), and Phantom Optimization (PO). If we do not use WA, we then use simple element-wise summation or averaging. In this case, we pick the better performances. Note that, PO-Step1 and -Step2 mean PO is applied in Step1 or Step2.}
\label{tab:4}
\vspace{-3mm}
\resizebox{\linewidth}{!}{
\renewcommand{\tabcolsep}{0.5mm}
\begin{tabular}{lccccccccccccc}
\toprule
& WA      & PD       & PO-Step1     & PO-Step2      & CV-Bench & BLINK & MMB  & SEED-Bench-2-Plus & VisualWebBench & MM-Vet & MM-Vet-v2 & LLaVA-Wilder & MathVista \\
\midrule
\multirow{7}{*}{\rotatebox[origin=c]{90}{Phantom-0.5B}}
& \xmark  & \xmark  & \xmark        & \xmark        & 28.2     & 21.4  & 60.7 & 35.7              & 34.7           & 26.6   & 22.0      & 60.8         & 33.8      \\
& \xmark  & \cmark  & \xmark        & \xmark        & 29.8     & 21.9  & 62.4 & 39.9              & 37.2           & 27.4   & 22.3      & 64.9         & 36.7      \\
& \cmark  & \cmark  & \xmark        & \xmark        & 38.1     & 27.4  & 70.1 & 43.7              & 42.3           & 31.8   & 29.7      & 69.7         & 40.0      \\
\cdashline{2-14}\noalign{\vskip 0.5ex}
& \cmark  & \cmark  & \cmark        & \xmark        & \textbf{41.5}& \textbf{39.3}& \textbf{72.7}& \textbf{47.9}& \textbf{51.8}& \textbf{45.7}& \textbf{41.5}& \textbf{72.2}& \textbf{51.7}\\
& \cmark  & \cmark  & \xmark        & \cmark        & 36.2     & 36.7  & 68.8 & 40.4              & 47.1           & 39.9   & \uu{36.6}      & 67.4         & \uu{48.2}      \\
& \cmark  & \cmark  & \cmark        & \cmark        & \uu{38.5}     & \uu{38.0}  & \uu{69.1} & \uu{45.5}              & \uu{47.2}           & \uu{42.3}   & 36.2      & \uu{71.0}         & 47.3\\
\cdashline{2-14}\noalign{\vskip 0.5ex}
& \xmark  & \xmark  & \cmark        & \xmark        & 32.0     & 24.2  & 64.2 & 39.0              & 36.4           & 31.2   & 24.1      & 63.4         & 36.8      \\
\midrule
\multirow{7}{*}{\rotatebox[origin=c]{90}{Phantom-1.8B}}
& \xmark  & \xmark  & \xmark        & \xmark        & 44.7     & 28.9  & 60.2 & 43.3              & 45.4           & 35.1   & 26.1      & 63.2         & 42.3      \\
& \xmark  & \cmark  & \xmark        & \xmark        & 47.0     & 32.6  & 64.7 & 44.9              & 46.5           & 36.0   & 27.4      & 68.7         & 46.4      \\
& \cmark  & \cmark  & \xmark        & \xmark        & 52.6     & 35.2  & 69.8 & 50.0              & 53.5           & 41.8   & 32.5      & 71.1         & 49.1      \\
\cdashline{2-14}\noalign{\vskip 0.5ex}
& \cmark  & \cmark  & \cmark        & \xmark        & \textbf{63.1}& \textbf{44.2}& \textbf{76.6}& \textbf{57.1}& \textbf{55.9}& \textbf{54.1}& \textbf{46.3}& \textbf{78.5}& \textbf{60.9}\\
& \cmark  & \cmark  & \xmark        & \cmark        & \uu{59.9}     & 39.9  & 72.2 & 49.7              & 48.4           & 50.5   & 37.0      & \uu{77.1}         & 55.9      \\
& \cmark  & \cmark  & \cmark        & \cmark        & 59.6     & \uu{40.6}  & \uu{73.7} & \uu{54.5}              & \uu{55.2}           & \uu{53.3}   & \uu{41.7}      & 76.0         & \uu{58.8}\\
\cdashline{2-14}\noalign{\vskip 0.5ex}
& \xmark  & \xmark  & \cmark        & \xmark        & 48.2     & 30.7  & 61.2 & 47.0              & 49.7           & 37.1   & 29.5      & 68.2         & 44.4      \\
\midrule
\multirow{7}{*}{\rotatebox[origin=c]{90}{Phantom-3.8B}}
& \xmark  & \xmark  & \xmark        & \xmark        & 63.7     & 34.4  & 62.6 & 42.9              & 45.6           & 38.1   & 32.6      & 73.5         & 45.3      \\
& \xmark  & \cmark  & \xmark        & \xmark        & 66.6     & 37.9  & 65.9 & 44.5              & 46.1           & 40.9   & 34.1      & 78.0         & 49.8      \\
& \cmark  & \cmark  & \xmark        & \xmark        & 69.1     & 44.1  & 68.9 & 51.8              & 51.8           & 46.9   & 37.0      & 83.4         & 50.5      \\
\cdashline{2-14}\noalign{\vskip 0.5ex}
& \cmark  & \cmark  & \cmark        & \xmark        & \textbf{73.8}& \textbf{51.5}& \textbf{80.4}& \textbf{61.8}& \textbf{59.8}& \textbf{54.4}& \textbf{48.5}& \textbf{85.7}& \textbf{60.6}\\
& \cmark  & \cmark  & \xmark        & \cmark        & 67.9     & 45.9  & 76.2 & 54.4              & \uu{56.8}           & \uu{50.6}   & 42.0      & 84.5         & 53.7      \\
& \cmark  & \cmark  & \cmark        & \cmark        & \uu{69.2}     & \uu{47.8}  & \uu{79.2} & \uu{58.6}              & 54.8           & 49.9   & \uu{42.9}      & \uu{85.0}         & \uu{58.1}\\
\cdashline{2-14}\noalign{\vskip 0.5ex}
& \xmark  & \xmark  & \cmark        & \xmark        & 68.6     & 37.6  & 65.5 & 47.0              & 46.8           & 41.3   & 36.3      & 76.1         & 49.7      \\
\midrule
\multirow{7}{*}{\rotatebox[origin=c]{90}{Phantom-7B}}
& \xmark  & \xmark  & \xmark        & \xmark        & 59.1     & 41.9  & 71.9 & 50.2              & 51.9           & 50.2   & 44.2      & 69.5         & 56.2      \\
& \xmark  & \cmark  & \xmark        & \xmark        & 59.8     & 45.9  & 72.5 & 54.2              & 53.6           & 53.7   & 46.3      & 74.4         & 60.9      \\
& \cmark  & \cmark  & \xmark        & \xmark        & 69.0     & 47.7  & 81.7 & 59.3              & 57.1           & 62.1   & 53.2      & 77.2         & 64.5      \\
\cdashline{2-14}\noalign{\vskip 0.5ex}
& \cmark  & \cmark  & \cmark        & \xmark        & \textbf{74.9}& \textbf{58.9}& \textbf{84.8}& \textbf{65.5}& \textbf{67.7}& \textbf{70.8}& \textbf{60.6}& \textbf{82.9}& \textbf{70.9}\\
& \cmark  & \cmark  & \xmark        & \cmark        & \uu{71.7}     & 52.2  & 77.9 & 59.1              & 64.1           & 68.2   & 53.1      & 78.7         & 68.1      \\
& \cmark  & \cmark  & \cmark        & \cmark        & 70.8     & \uu{54.4}  & \uu{82.9} & \uu{60.5}              & \uu{66.6}           & \uu{69.0}   & \uu{54.5}      & \uu{82.6}         & \uu{68.3}\\
\cdashline{2-14}\noalign{\vskip 0.5ex}
& \xmark  & \xmark  & \cmark        & \xmark        & 61.1     & 43.0  & 75.9 & 52.4              & 53.4           & 54.8   & 47.1      & 73.3         & 59.3      \\
\bottomrule
\end{tabular}
}
\vspace{-7mm}
\end{table}

%% file: appendix/AppendixA.tex
\begin{figure}[h!]
    \centering
    \includegraphics[width=\textwidth]{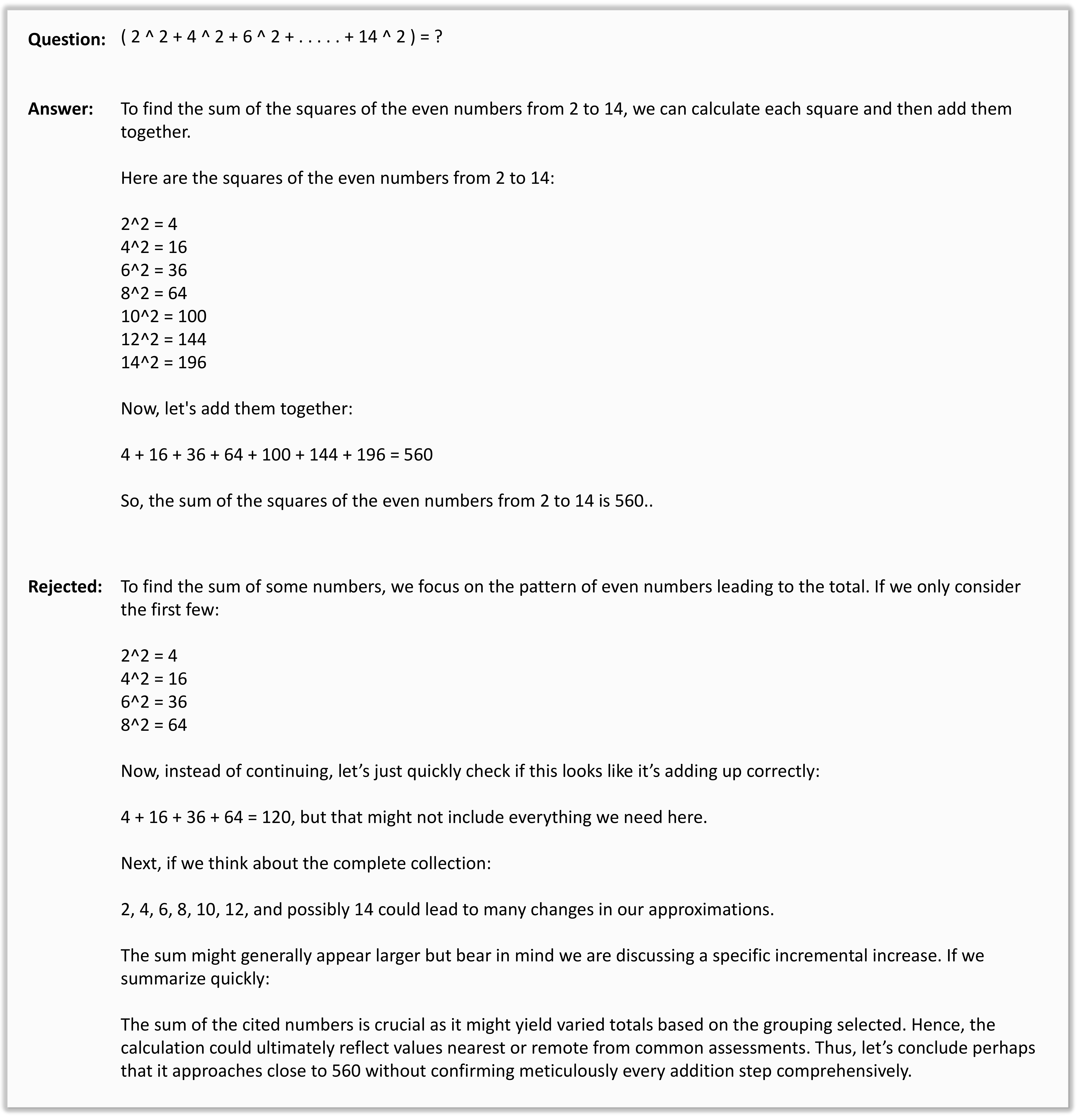}
\end{figure}

\begin{figure}[h!]
    \centering
    \includegraphics[width=\textwidth]{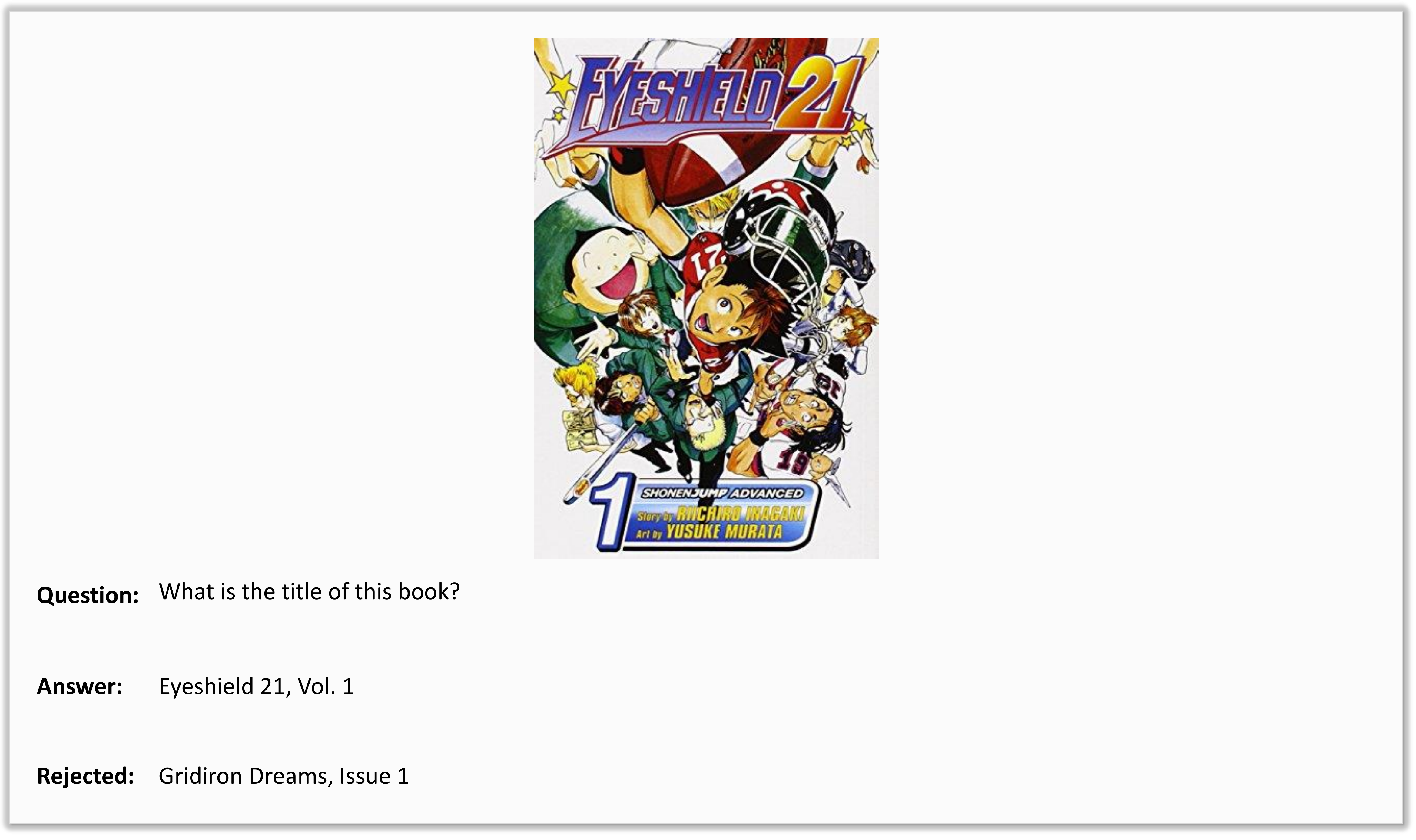}
\end{figure}

\begin{figure}[h!]
    \centering
    \includegraphics[width=\textwidth]{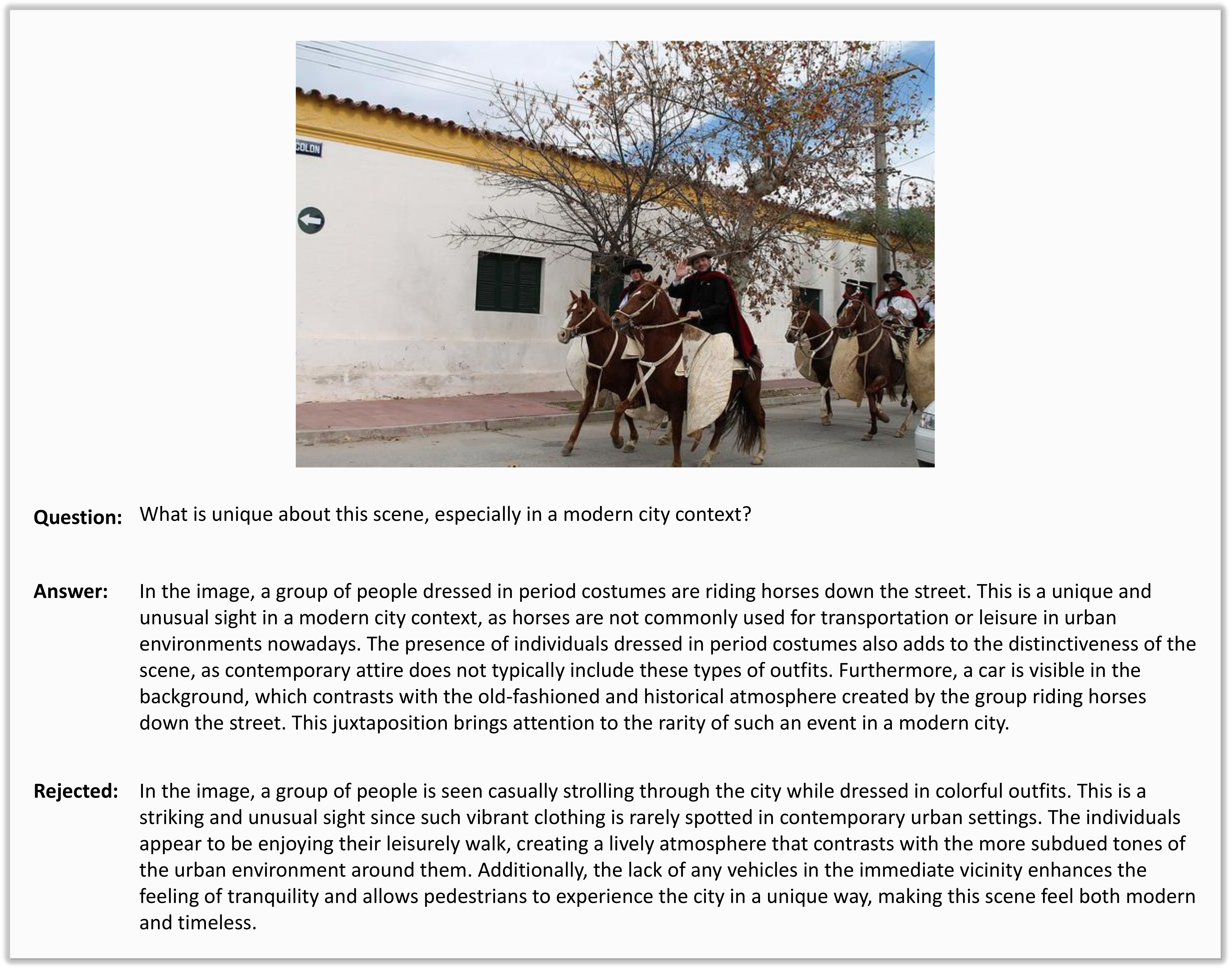}
\end{figure}

\begin{figure}[h!]
    \centering
    \includegraphics[width=\textwidth]{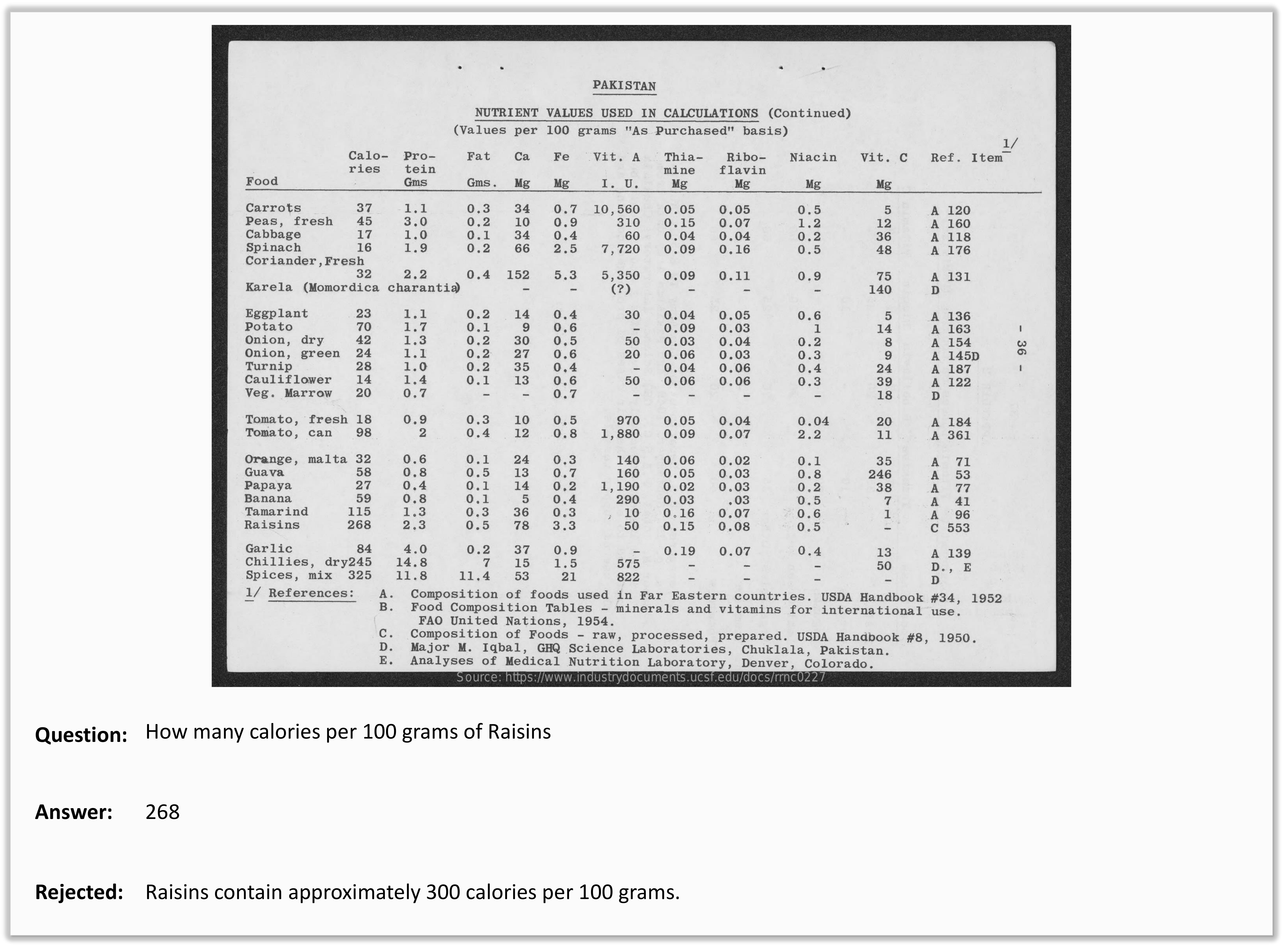}
\end{figure}

\begin{figure}[h!]
    \centering
    \includegraphics[width=\textwidth]{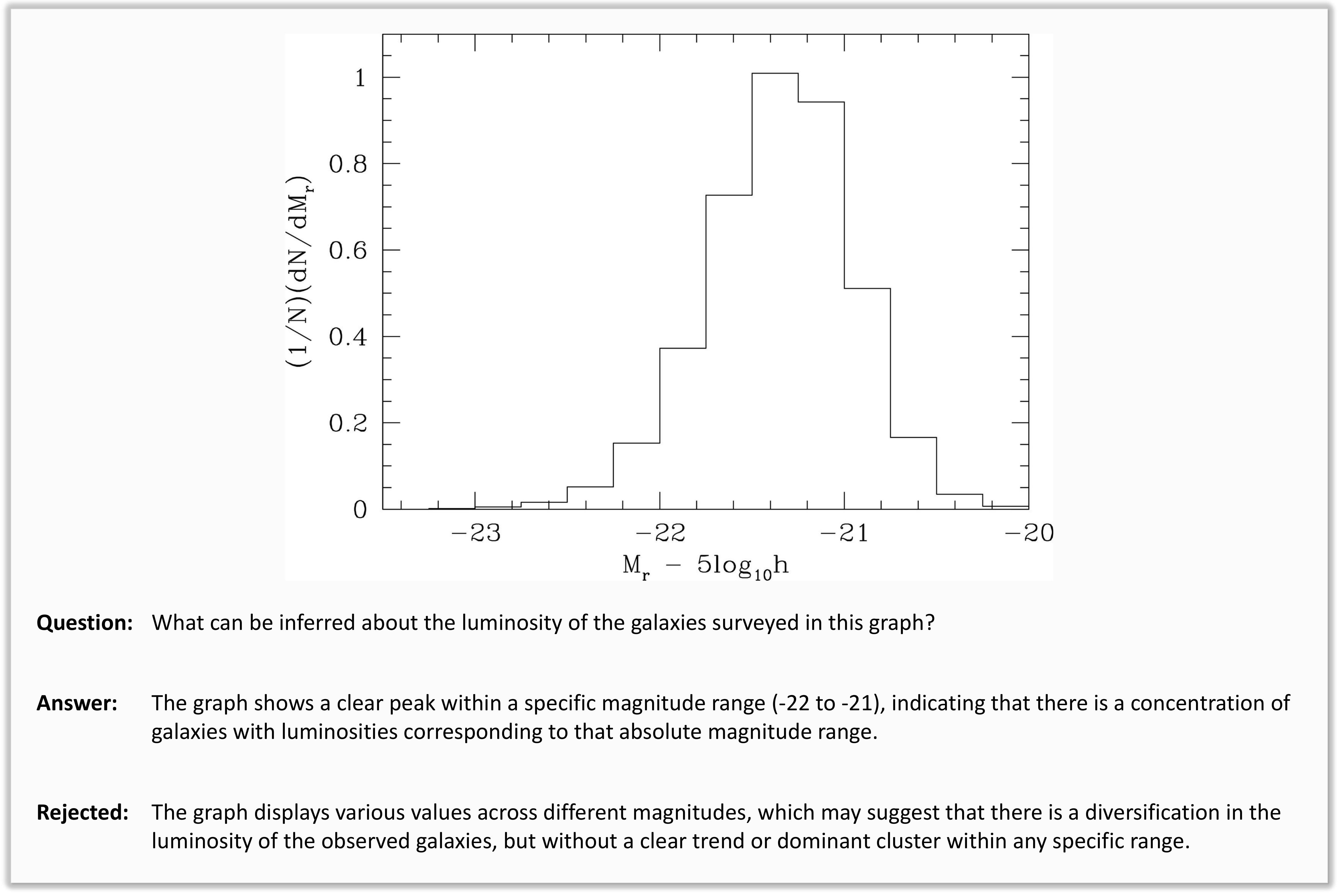}
\end{figure}

\begin{figure}[h!]
    \centering
    \includegraphics[width=\textwidth]{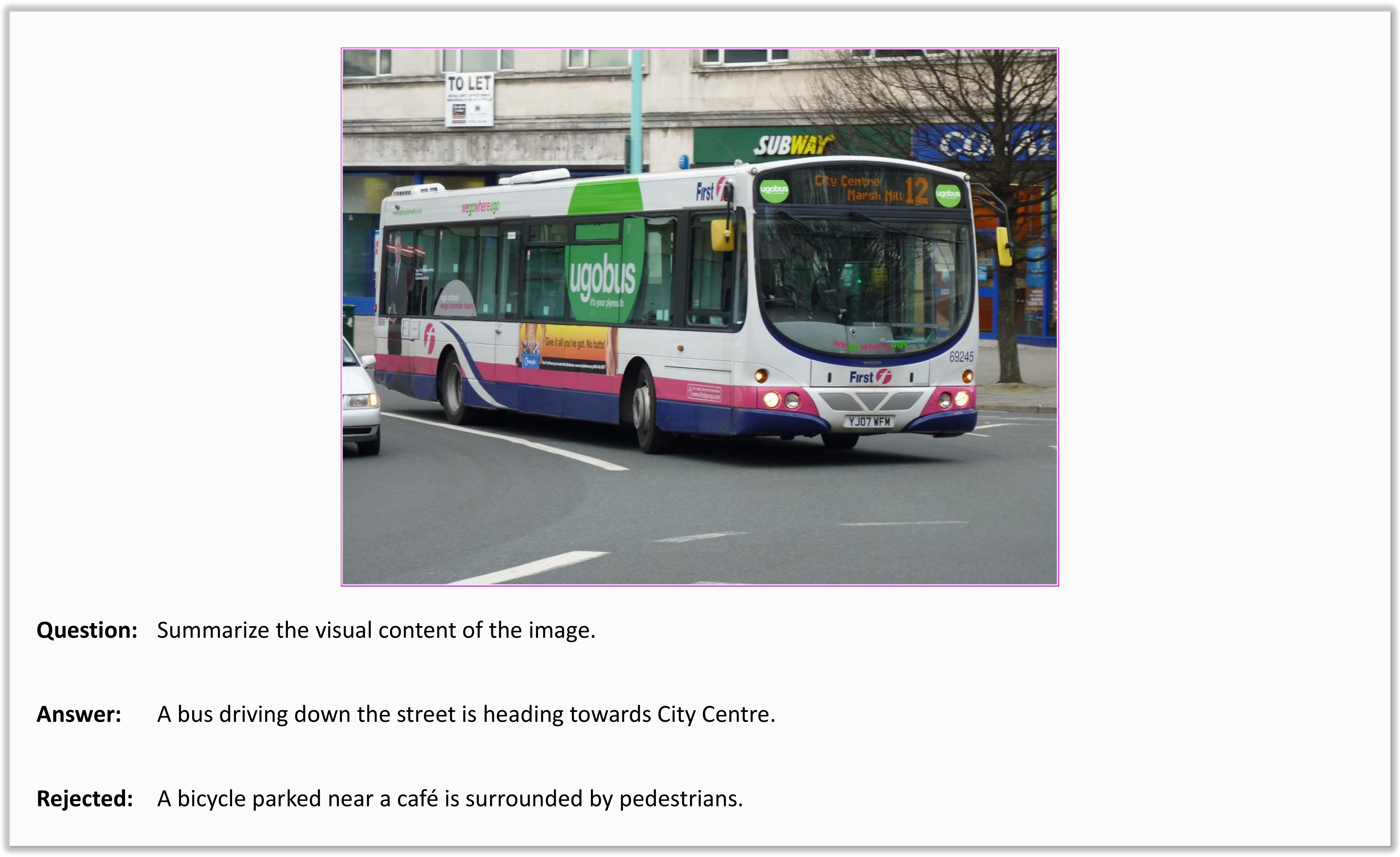}
\end{figure}

\begin{figure}[h!]
    \centering
    \includegraphics[width=\textwidth]{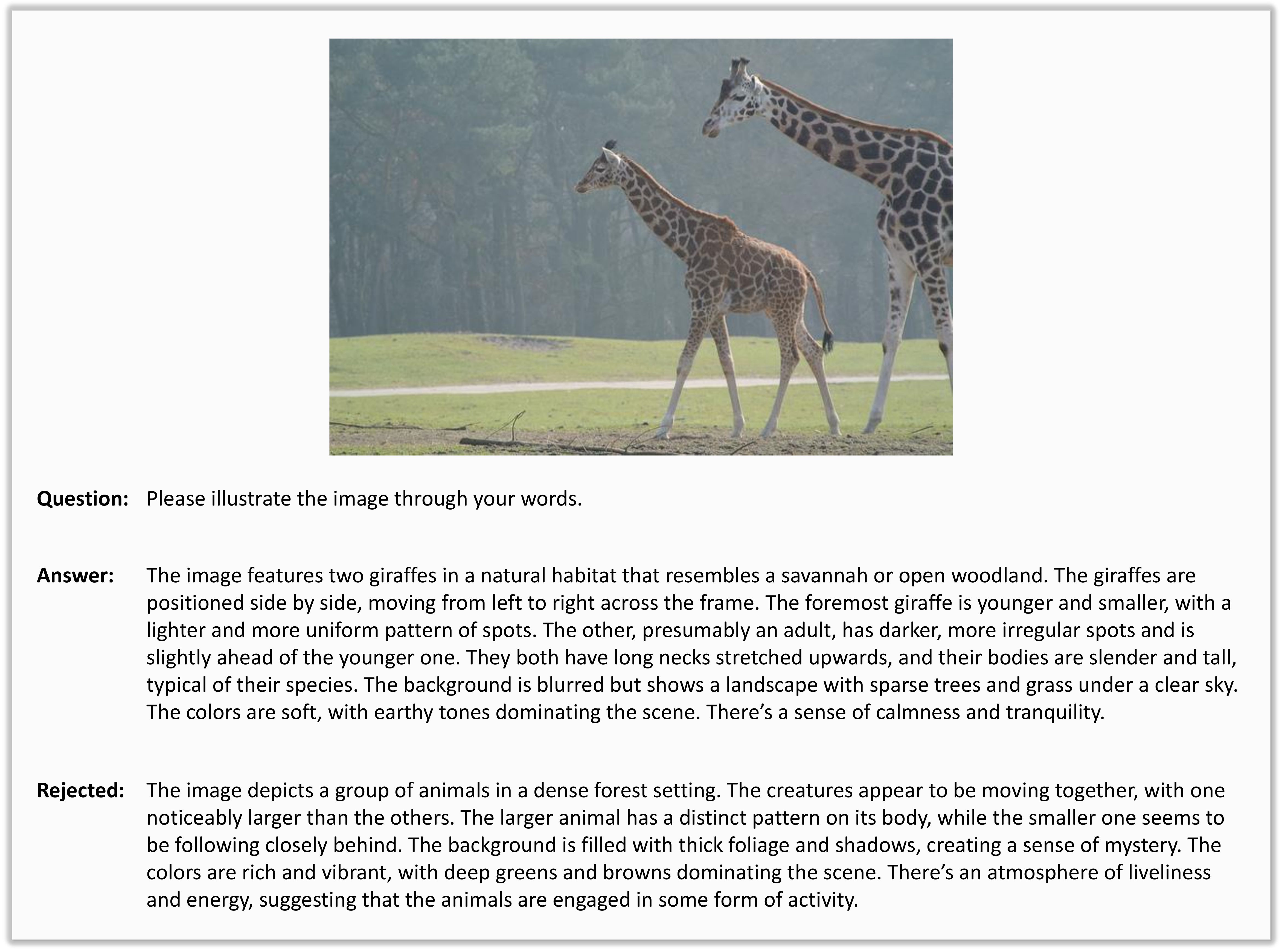}
\end{figure}

%% file: appendix/AppendixC.tex
\begin{figure}[h!]
    \centering
    \includegraphics[width=\textwidth]{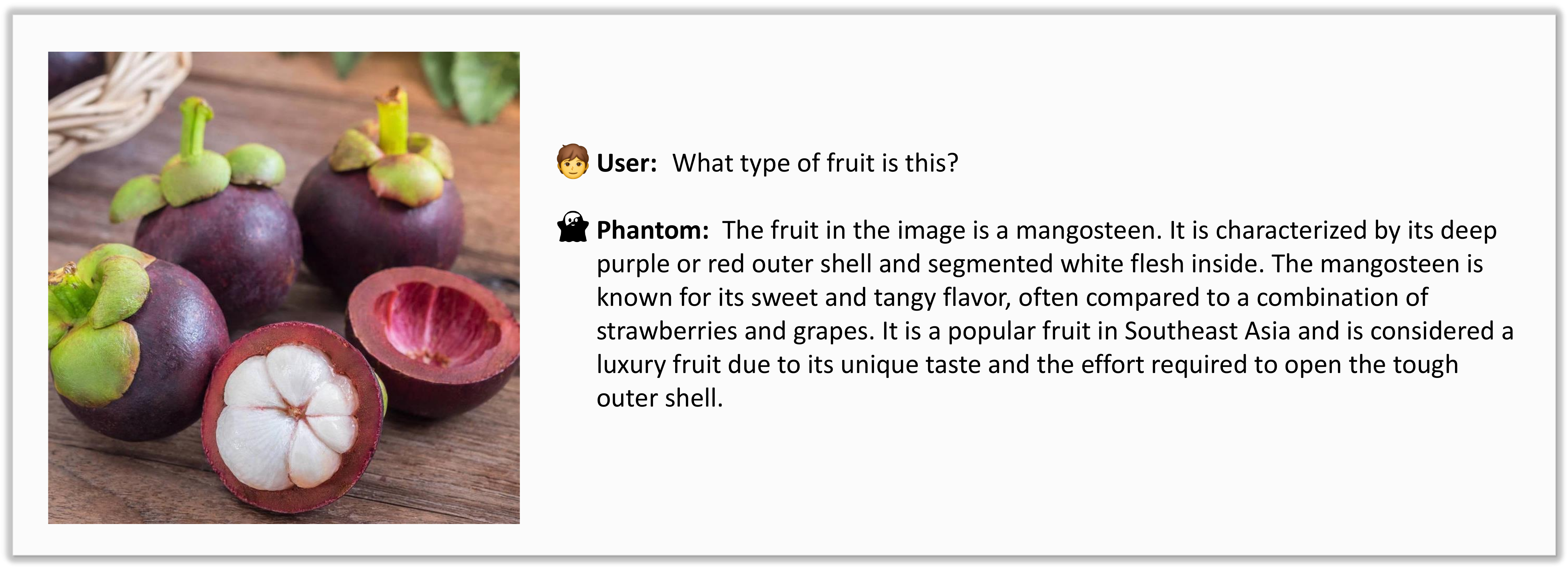}
\end{figure}

\begin{figure}[h!]
    \centering
    \includegraphics[width=\textwidth]{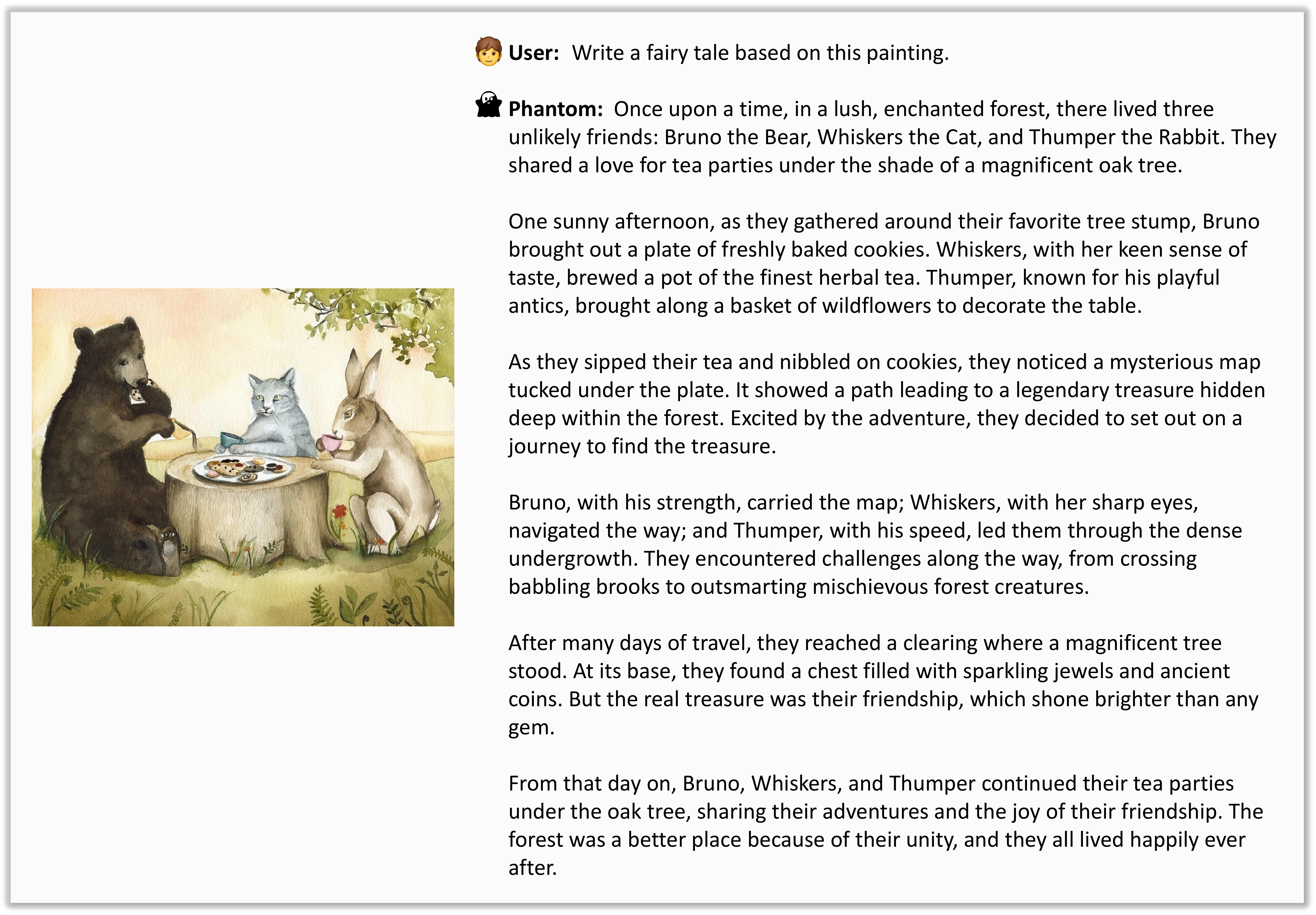}
\end{figure}

\begin{figure}[h!]
    \centering
    \includegraphics[width=\textwidth]{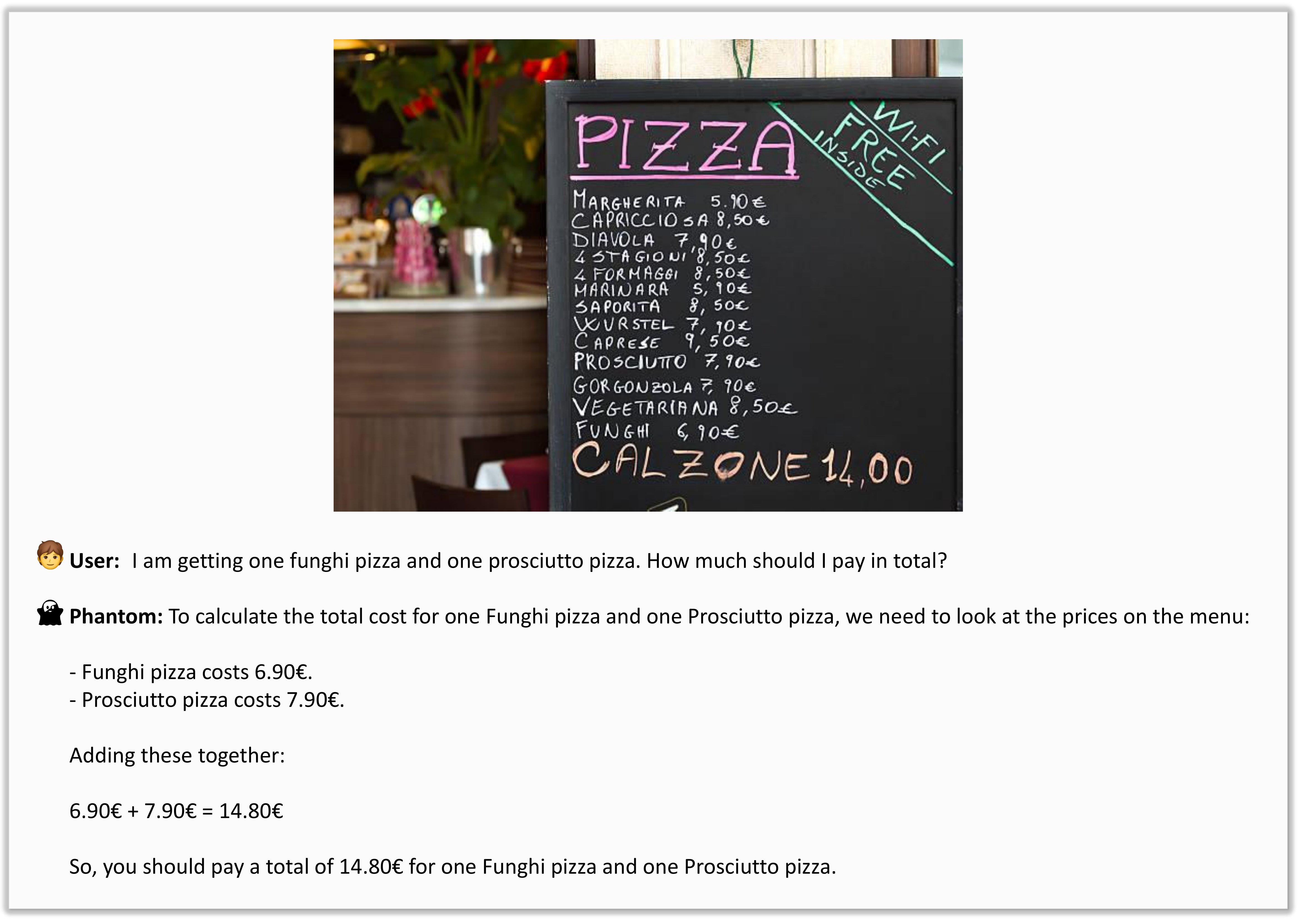}
\end{figure}

\begin{figure}[h!]
    \centering
    \includegraphics[width=\textwidth]{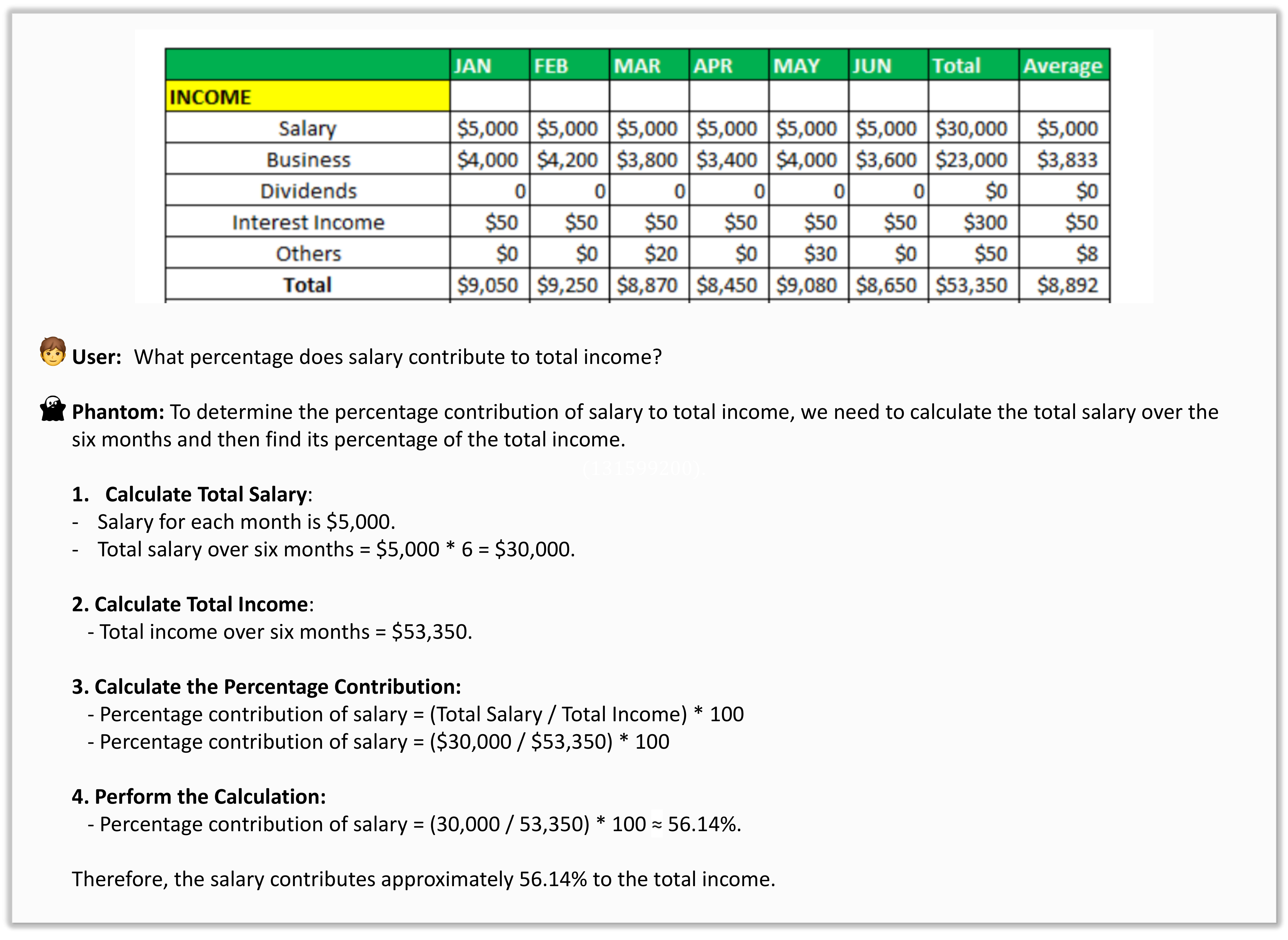}
\end{figure}

\begin{figure}[h!]
    \centering
    \includegraphics[width=\textwidth]{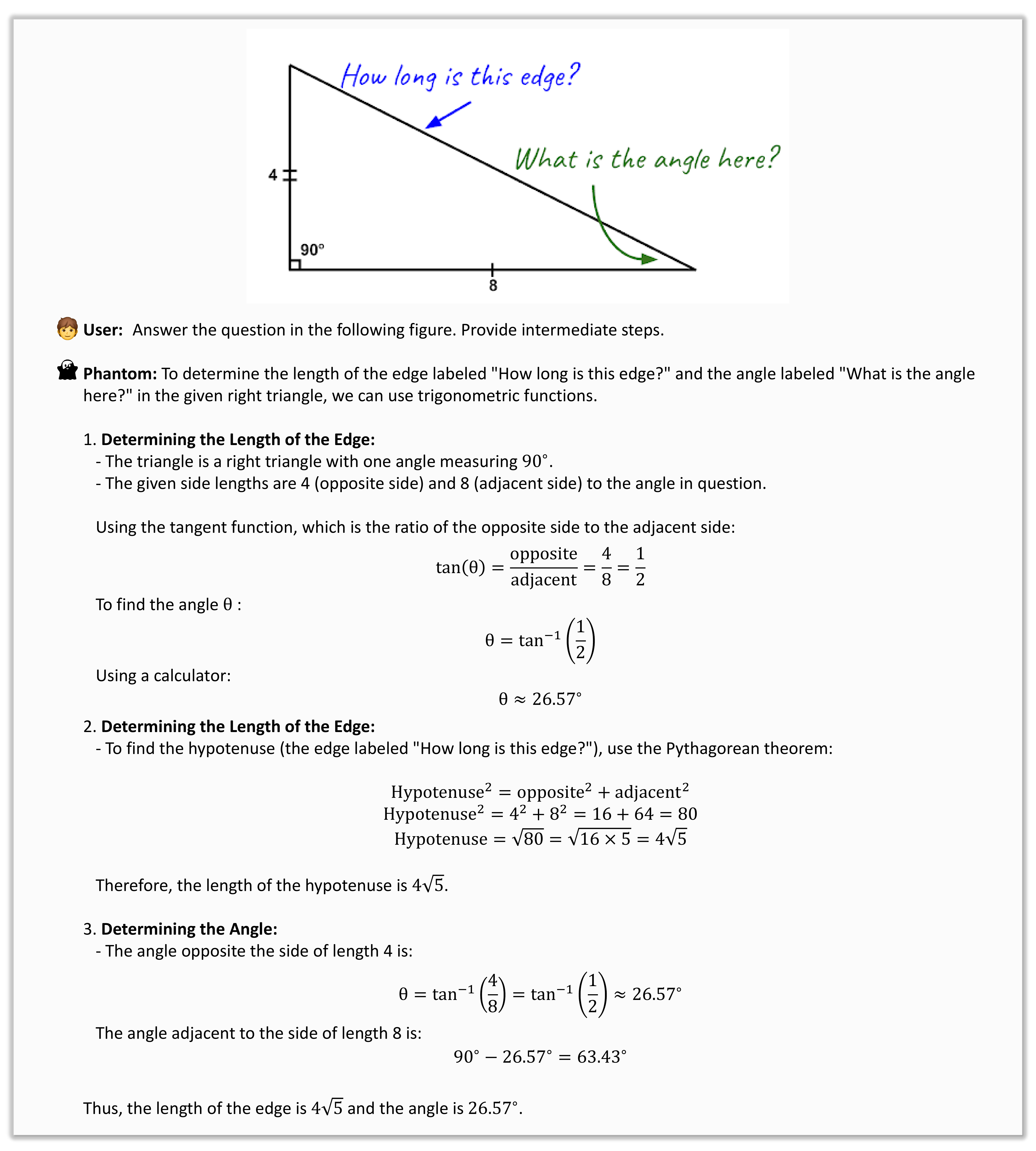}
\end{figure}

\begin{figure}[h!]
    \centering
    \includegraphics[width=\textwidth]{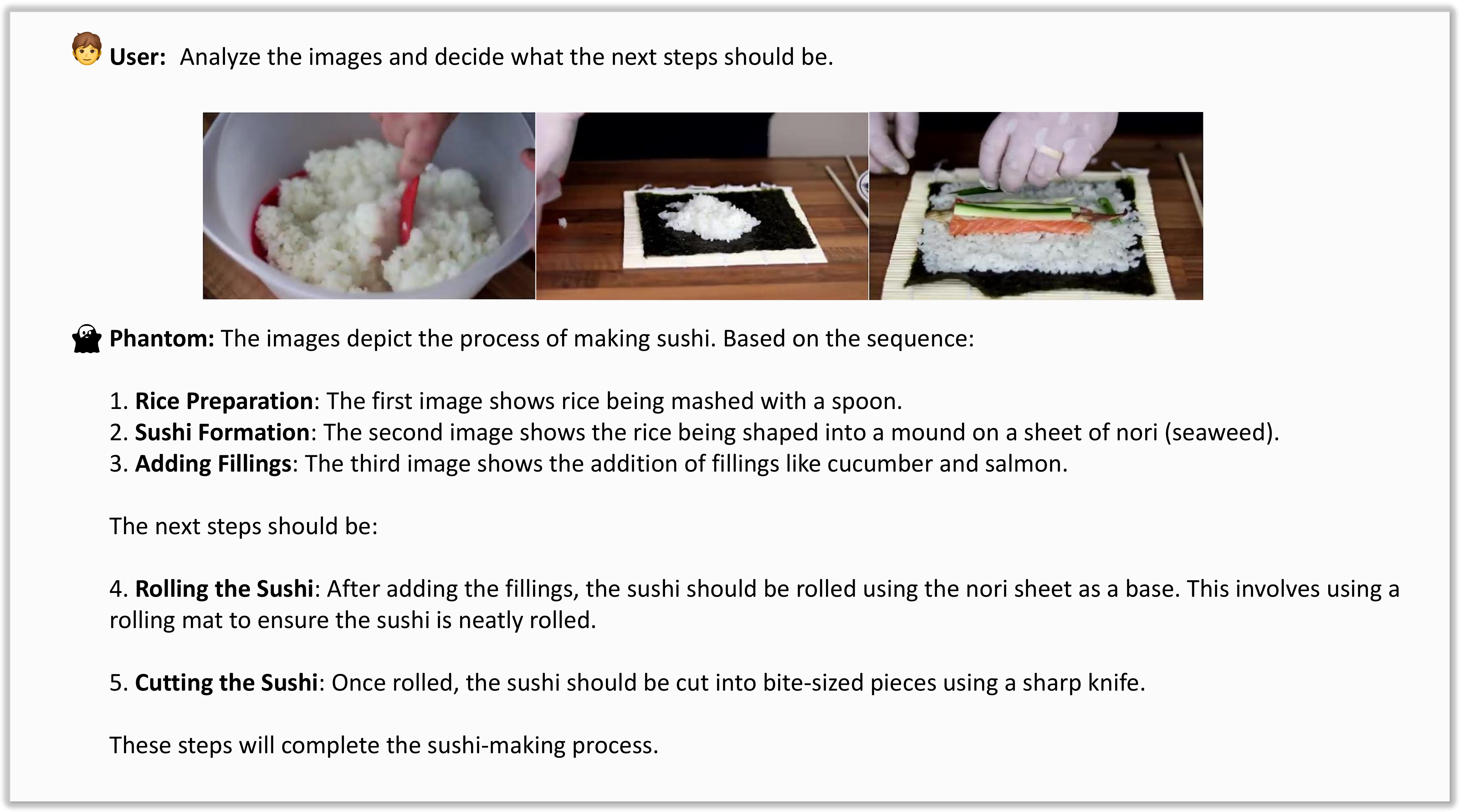}
\end{figure}

\begin{figure}[h!]
    \centering
    \includegraphics[width=\textwidth]{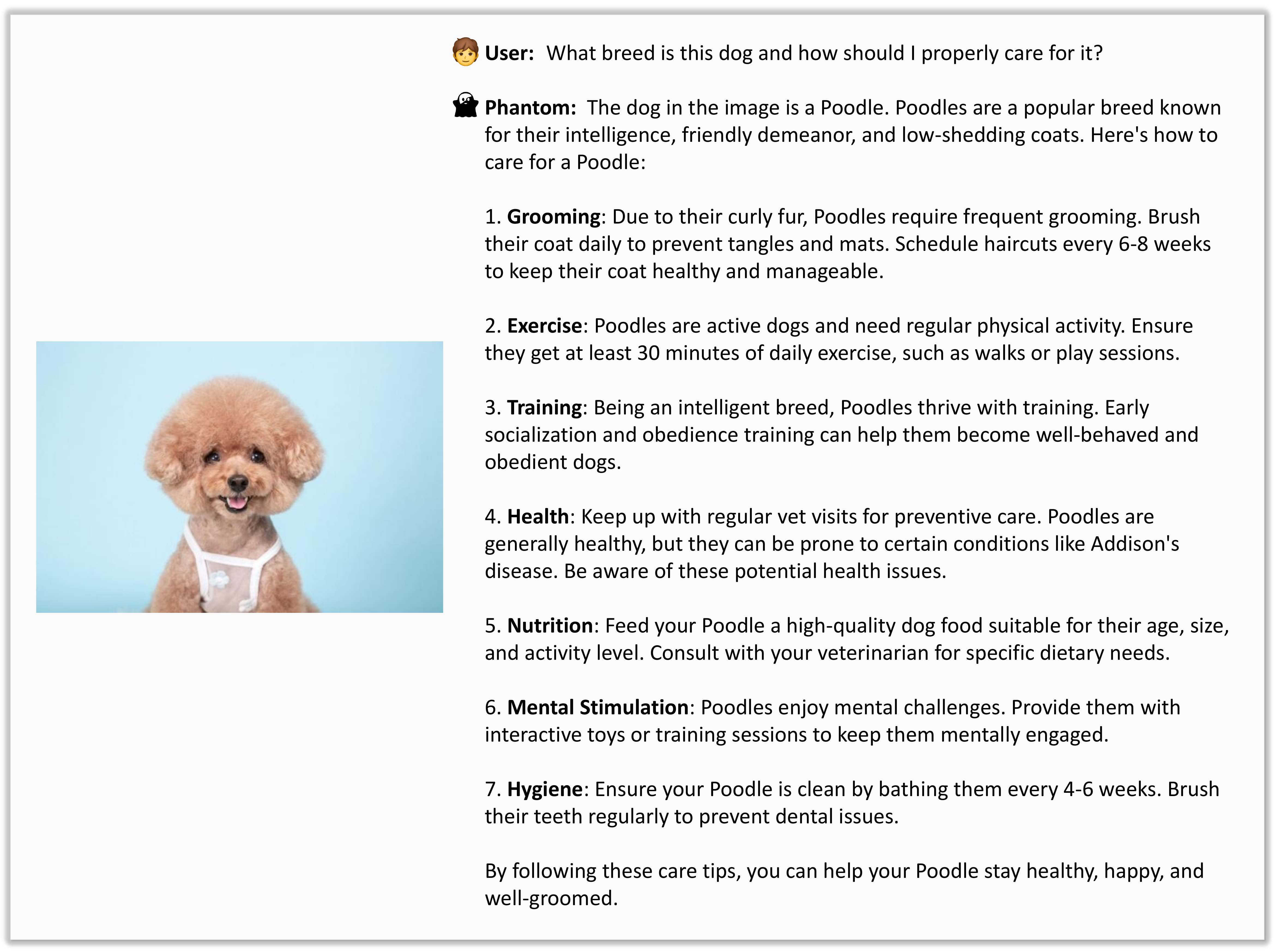}
\end{figure}

\begin{figure}[h!]
    \centering
    \includegraphics[width=\textwidth]{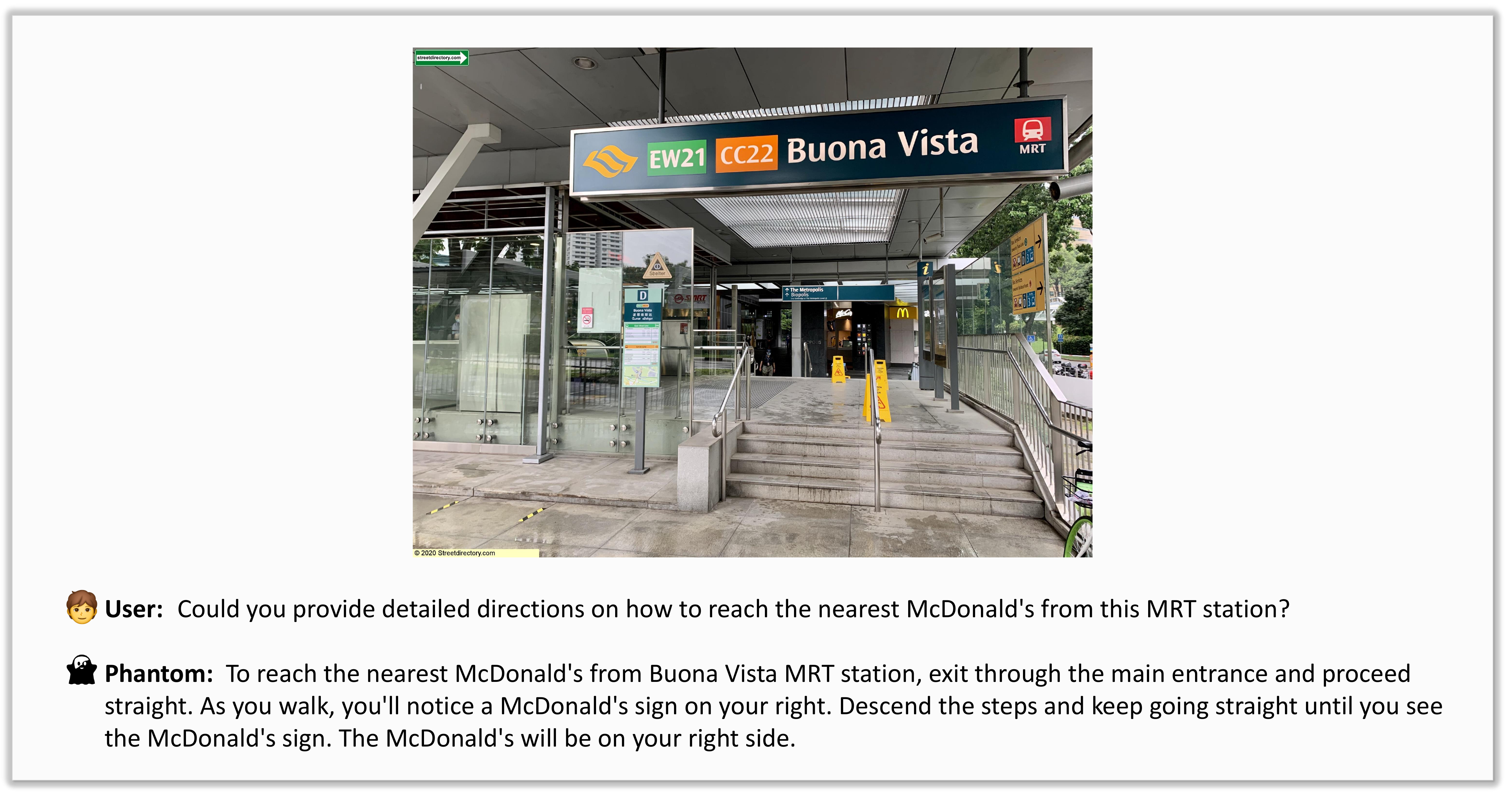}
\end{figure}